\documentclass[10pt,twocolumn,letterpaper]{article}
\usepackage{iccv}              
\usepackage[linesnumbered,ruled,vlined]{algorithm2e}
\usepackage{tabularx} 
\usepackage{color,colortbl} 
\usepackage{multirow}
\usepackage{float}
\usepackage{subcaption}
\usepackage[accsupp]{axessibility}  

\usepackage{amsmath, amssymb, bm}
\usepackage{tikz}

\definecolor{iccvblue}{rgb}{0.21,0.49,0.74}
\usepackage[pagebackref,breaklinks,colorlinks,allcolors=iccvblue]{hyperref}

\def\confName{ICCV}
\def\confYear{2025}

\title{Robust 3D-Masked Part-level Editing in 3D Gaussian Splatting \\with Regularized Score Distillation Sampling}

\author{
Hayeon Kim$^{1,*}$ \qquad
Ji Ha Jang$^{1,*}$ \qquad
Se Young Chun$^{1,2,\dagger}$ \\
$^1$ Dept. of Electrical and Computer Engineering, $^2$ INMC \& IPAI \\
Seoul National University, Republic of Korea \\
{\tt\small \{khy5630, jeeit17, sychun\}@snu.ac.kr}
}
\footnotetext{Authors contributed equally. $^{\dagger}$ Corresponding author.}

\begin{document}
\begin{figure*}
\twocolumn[{
    \maketitle
    \begin{center}
    \includegraphics[width=\textwidth]{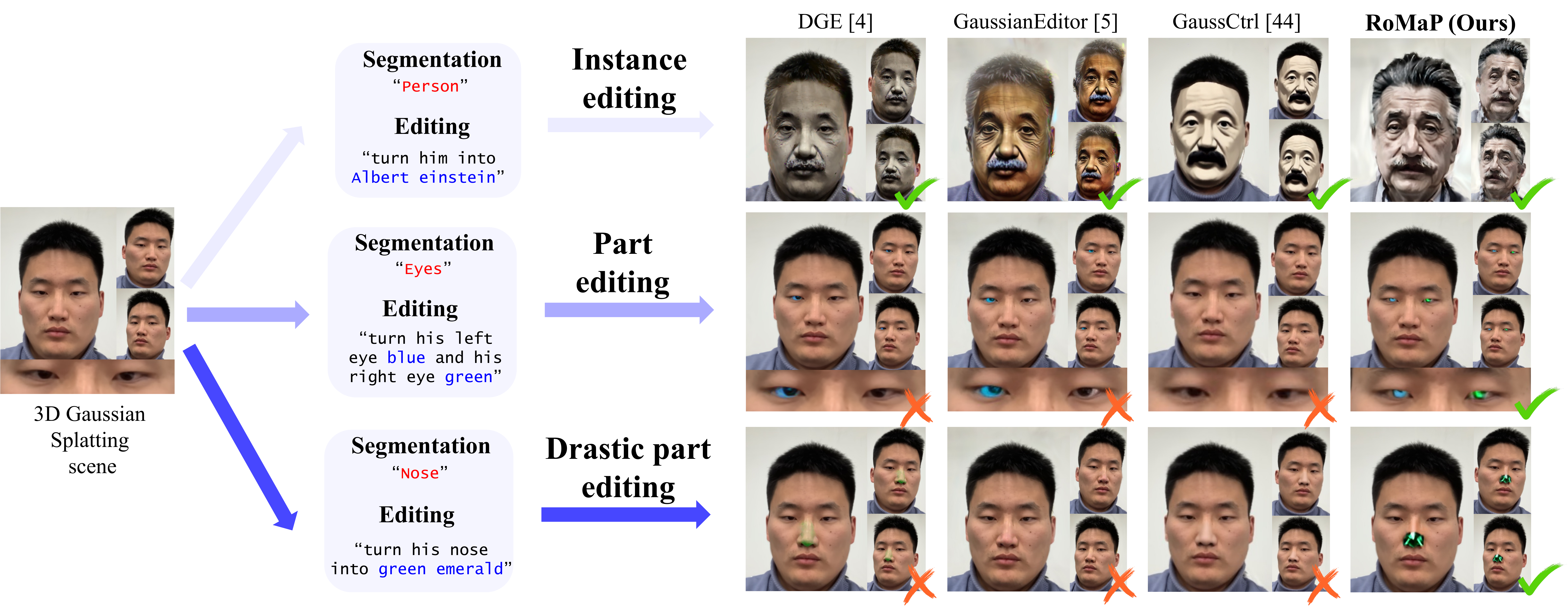}
    \vspace{-1cm}
    \newline
    \caption{
    \textbf{Enhanced controllability in 3D Gaussian part-level editing achieved with RoMaP, surpassing prior arts.}
    RoMaP enables highly controllable and localized part-level edits, allowing even for unconventional modifications such as `emerald nose' or modifications requiring a high-level controllability such as `blue left eye, right green eye' while maintaining global consistency. In contrast, existing baselines perform well for instance-level editing, but struggle with part-level editing, especially with drastic changes.}
    \label{fig:front_results}
    \end{center}
}]

\end{figure*}

\begin{abstract}
Recent advances in 3D neural representations and instance-level editing models have enabled the efficient creation of high-quality 3D content. However, achieving precise local 3D edits remains challenging, especially for Gaussian Splatting, due to inconsistent multi-view 2D part segmentations and inherently ambiguous nature of Score Distillation Sampling (SDS) loss. To address these limitations, we propose RoMaP, a novel local 3D Gaussian editing framework that enables precise and drastic part-level modifications.
First, we introduce a robust 3D mask generation module with our 3D-Geometry Aware Label Prediction (3D-GALP), which uses spherical harmonics (SH) coefficients to model view-dependent label variations and soft-label property, yielding accurate and consistent part segmentations across viewpoints.
Second, we propose a regularized SDS loss that combines the standard SDS loss with additional regularizers. In particular, an $\mathcal{L}_1$ anchor loss is introduced via our Scheduled Latent Mixing and Part (SLaMP) editing method, which generates high-quality part-edited 2D images and confines modifications only to the target region while preserving contextual coherence. Additional regularizers, such as Gaussian prior removal, further improve flexibility by allowing changes beyond the existing context, and robust 3D masking prevents unintended edits.
Experimental results demonstrate that our RoMaP achieves state-of-the-art local 3D editing on both reconstructed and generated Gaussian scenes and objects qualitatively and quantitatively, making it possible for more robust and flexible part-level 3D Gaussian editing.  Code is available at \href{https://janeyeon.github.io/romap}{https://janeyeon.github.io/romap}.
\end{abstract}    
\section{Introduction}
\label{sec:intro}

\begin{figure*}
    \centering
     \includegraphics[width=\textwidth]{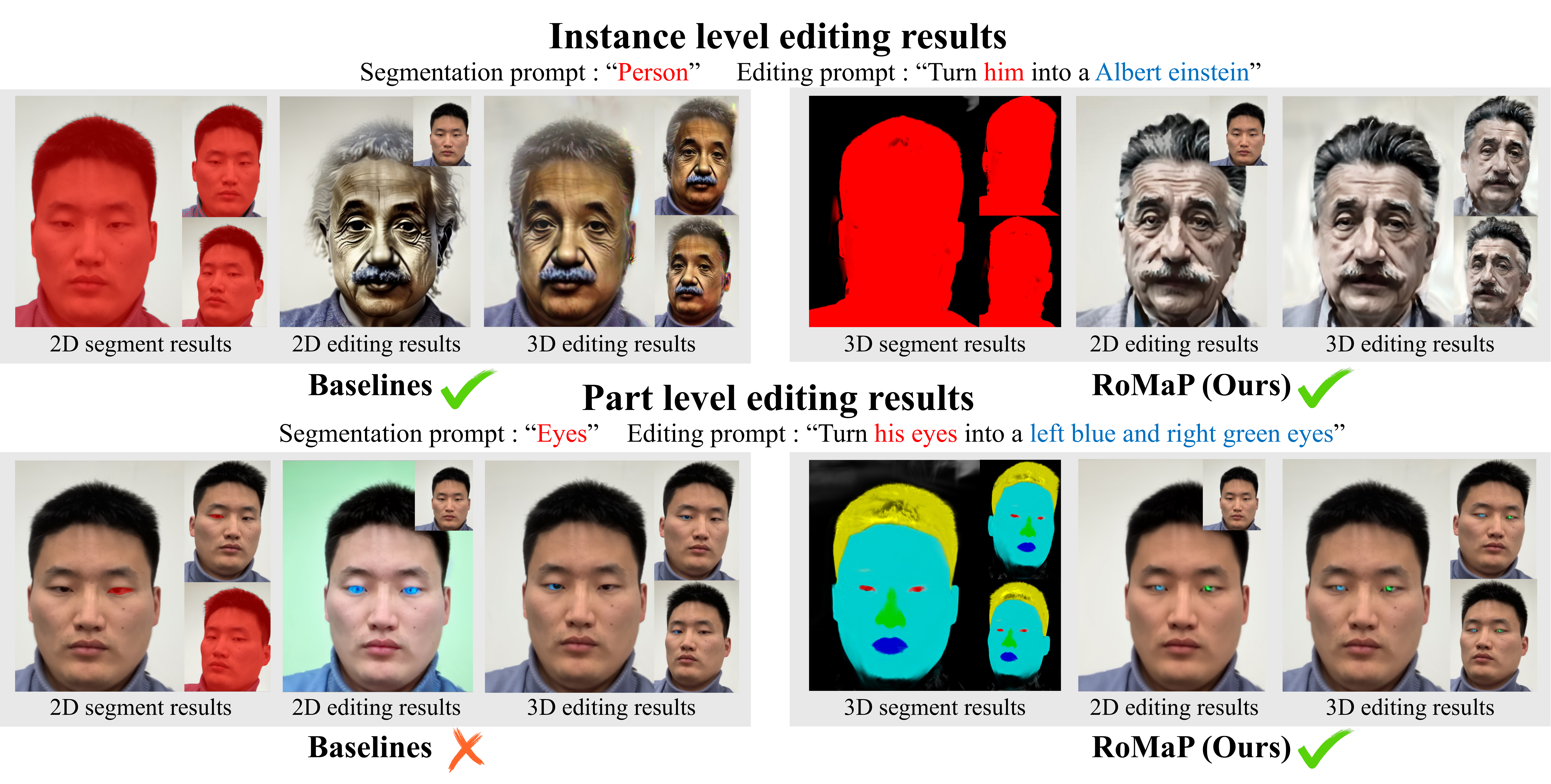}
     \vspace{-0.8cm}
    \caption{\textbf{Limitations of prior local 3D editing methods leveraging 2D part level segmentation and edits.} Although existing 3D editing methods excel in instance level editing, they struggle with part level editing as part segmentation~\cite{kirillov2023segment} (for `eye') lacks view consistency, and 2D editing~\cite{brooks2023instructpix2pix} often misplaces changes, turning a wall green instead of the left eye. In contrast, our method achieves accurate 3D eye segmentation with geometric awareness and clearly defines modification direction, enabling successful 3D Gaussian editing.
}
    \label{fig:challenge}
\end{figure*}

Recent advances in 3D neural representations~\cite{mildenhall2021nerf, kerbl20233d, wang2021neus, yariv2021volume, shen2021deep} and generative models~\cite{sohl2015deep, ho2020denoising} have enabled efficient, high-quality 3D content creation, increasingly vital for industries such as mixed reality and robotics. Unlike traditional, labor-intensive methods, text-to-image diffusion models~\cite{rombach2022high, ramesh2022hierarchical, ding2021cogview, luo2023latent} generate contents from text prompts, potentially reducing production costs and effort significantly. Enhancing controllability in 3D content generation is crucial for customizing these assets. Text-guided editing methods~\cite{palandra2024gsedit, chen2024gaussianeditor,chen2024dge,wu2024gaussctrl, cheng2023progressive3d, 10.1145/3664647.3681039} enhance this by enabling flexible expression of abstract and specific concepts while enabling edits at various levels of detail. 

Local 3D editing involves modifying part-level attributes like texture and color, or replacing parts. While previous works~\cite{haque2023instruct, dong2023vica, cheng2023progressive3d, chen2024gaussianeditor, wang2024gaussianeditor, chen2024dge, wu2024gaussctrl} have achieved excellent performance in instance-level 3D editing, local 3D editing remains challenging~(See Fig.~\ref{fig:front_results}). Prior methods~\cite{chen2024gaussianeditor,wang2024gaussianeditor,chen2024dge,wu2024gaussctrl} often use 2D segmentation~\cite{kirillov2023segment} to localize changes and apply 2D multi-view editing~\cite{brooks2023instructpix2pix} for 3D modifications. However, these approaches face two major challenges for part-level modifications, often leading to inaccurate or no edit. 

First, achieving consistent 3D editing across multiple views requires precise masking to preserve unchanged regions, typically relying on 2D multi-view image segmentation. However, compared to instance segmentation, part segmentation is challenging due to occlusions and variations in appearance across viewpoints. Existing approaches~\cite{chen2024gaussianeditor,wang2024gaussianeditor,chen2024dge,wu2024gaussctrl} leverage language-based SAM~\cite{kirillov2023segment} to segment target parts in multi-view images and re-project them onto 3D for editing. While 2D instance segmentation remains consistent across views, part-level segmentation is much less reliable (\textit{e.g.,} some views may capture only one eye, merge both, or miss them entirely), resulting in unstable and incomplete masks, as shown in Fig.~\ref{fig:challenge}. Additionally, assigning a hard segmentation label to each Gaussian from a 2D map may be inappropriate, as Gaussians at part boundaries could represent different parts depending on the view, thus resulting in mixed soft-labels.

Second, part-level 3D editing remains challenging as existing models struggle to isolate and modify specified parts~\cite{brooks2023instructpix2pix} or handle semantically low-probability edits~\cite{wang2023nerf}. Learned part-instance correlations often cause unintended changes or failures when the target attribute deviates from the original context. As shown in Fig.~\ref{fig:challenge}, InstructPix2Pix~\cite{brooks2023instructpix2pix}, widely used for 2D editing in prior works~\cite{chen2024dge, chen2024gaussianeditor, haque2023instruct, dong2023vica}, excels in instance edits but struggles with part edits. Instead of applying precise direct changes to the eyes, the model alters the background to green and turns the eyes blue, as odd-eye coloration is rare in human faces, making the edit statistically more likely. Moreover, achieving such fine-grained control remains highly challenging.

To address this challenge, we introduce RoMaP, a novel part-level 3D editing framework that enables precise and substantial local modifications for Gaussian. RoMaP comprises two core components:
(1) A robust 3D mask generation module with 3D-Geometry Aware Label Prediction (3D-GALP): 3D-GALP leverages spherical harmonics (SH) coefficients to explicitly model view-dependent label variations, effectively capturing the mixed-label property of Gaussians. This results in accurate and consistent part segmentations across viewpoints, enabling reliable local edits.
(2) A regularized Score Distillation Sampling (SDS) loss: Our regularized SDS combines the standard SDS loss with additional regularizers, including an $\mathcal{L}_1$ anchor loss from Scheduled Latent Mixing and Part (SLaMP) edited images. SLaMP generates 2D multi-view images with drastic changes strictly confined to the target region, guiding SDS optimization toward the intended modification. Additionally, robust 3D masking prevents unintended changes. Gaussian prior removal allows flexible adjustments, and together they enable precise local 3D editing, even along rare or unconventional directions.
Our RoMaP enables local 3D Gaussian editing, allowing diverse changes in specific areas. As seen in Fig.~\ref{fig:front_results}, our RoMaP achieved even drastic local edits, enabling unlikely or unconventional modifications while preserving the original identity, thereby enhancing controllability in 3D content editing. Our contributions are summarized as:
\begin{itemize}
\item Proposing RoMaP for precise and consistent local 3D Gaussian editing, enabled through our robust full 3D mask using our 3D-geometry aware label prediction, exploiting the uncertainty in soft-label Gaussians.
\item Proposing regularized SDS loss, enabling drastic part edits with scheduled latent mixing part editing and robust masks, along with Gaussians prior removal.
\item Experiments show that RoMaP enhances 3D Gaussian editing quality both qualitatively and quantitatively across reconstructed and generated Gaussian scenes and objects, improving controllability in 3D content generation.
\end{itemize}

\section{Related Works}
\label{sec:relatedworks}
\subsection{Diffusion and Rectified Flow based generation}
Recent advances in Diffusion Models (DMs)~\cite{rombach2022high, esser2024scaling} have greatly enhanced image generation, excelling in tasks like image editing, stylization~\cite{hertz2022prompt, kawar2023imagic, Wu_2023_ICCV, qi2024deadiff}. Rectified Flows (RFs)~\cite{liu2022flow}, a flow-based approach~\cite{dinh2016density}, streamline diffusion by linearizing the its path, enabling more efficient training, faster sampling, and more accurate latent space inversion. Recent combinations of RF and Diffusion Transformer (DiT)~\cite{peebles2023scalable} models, like FLUX and Stable Diffusion 3 (SD3)~\cite{esser2024scaling}, have advanced high-quality image generation, benefiting applications like text-to-3D and image editing. These models enhance prompt-faithful editing and inversion by refining noise~\cite{yang2024text} and utilizing RF's linearity~\cite{rout2024semantic} but still lack part-level controllability. Similarly, prior works~\cite{li2024dreambeast, li2024dreamcouple, yang2024text} employ these models in text-to-3D, achieving high-fidelity and faster convergence. Notably, SD3 has been applied to part-level controllable text-to-3D generation~\cite{li2024dreambeast} but is limited to animals, leaving broader applications unexplored. Our approach enables previously unattainable drastic local 3D edits by leveraging the SD3's part-awareness and RF's linearity, allowing flexible edits across various reconstructed and generated Gaussians.
\begin{figure*}[!ht]
    \centering
     \includegraphics[width=1.0\textwidth]{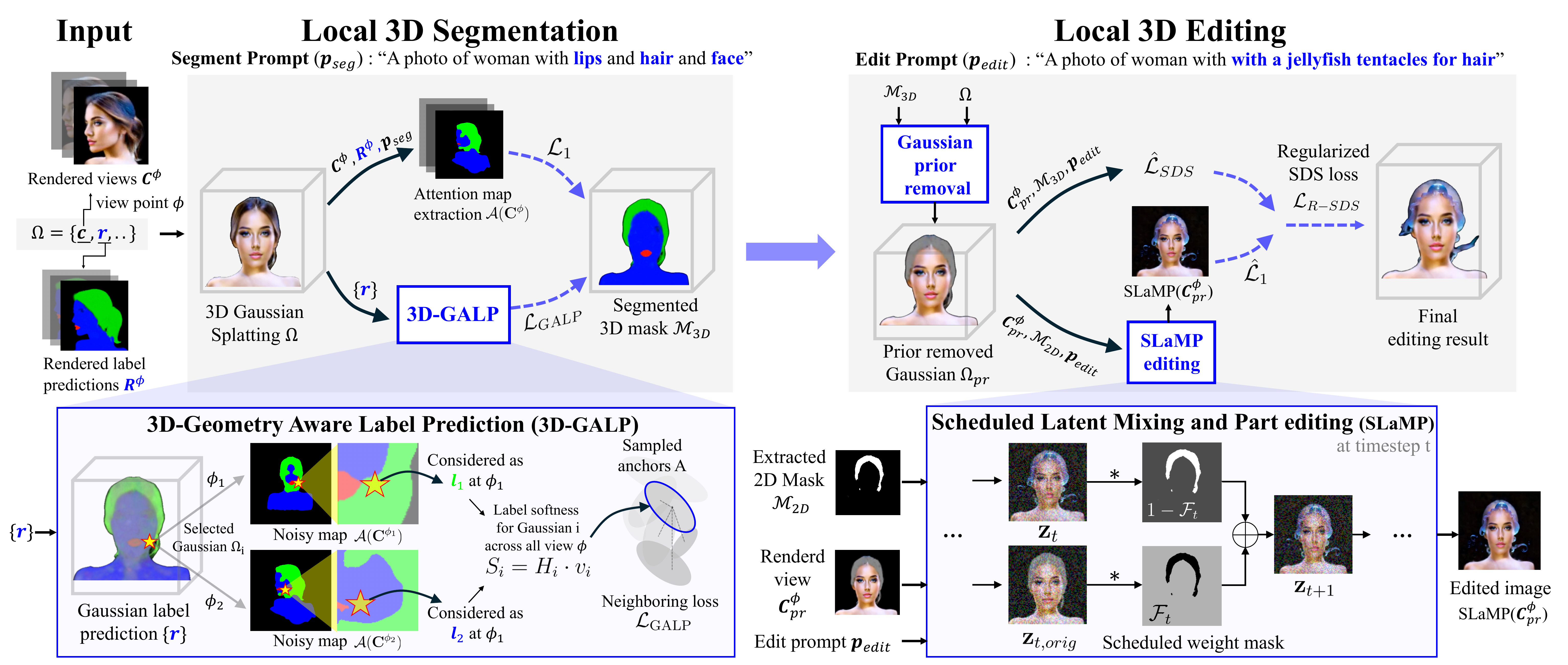}
    \caption{\textbf{Overall pipeline of RoMaP.} RoMaP first segments 3D Gaussian using 3D-GALP, leveraging the soft-label properties of Gaussians to address the intricacies of part-level segmentation. With anchors consisting of both label-consistent and inconsistent Gaussians, we refine 3D segmentation considering locality with neighboring Gaussians. Then, in local 3D editing, we first remove Gaussian priors and introduce a new modification direction using SLaMP-edited images, followed by refinement via a regularized SDS loss.
}
    \label{fig:pipeline}
\end{figure*}
\subsection{Editing of 3D Gaussian Splatting}
\label{Sec_2.2}
Editing 3D neural fields has advanced 3D generation by enhancing controllability, attracting research interest~\cite{haque2023instruct, li2024focaldreamer}. Early works focused on Neural Radiance Field (NeRF)~\cite{haque2023instruct, dong2023vica, Koo:2024PDS}, but recent works have shifted toward 3D Gaussians for better local control and efficient rendering. Editing Gaussians requires both an editing and a masking strategy to target specific parts. In editing strategy, some methods~\cite{chen2024gaussianeditor,chen2024dge,wu2024gaussctrl} edit 2D-rendered Gaussian images from multiple views using image editing models~\cite{brooks2023instructpix2pix, igs2gs, chen2024dge} and project them back onto Gaussians. However, this approach is limited by the constraints of the 2D editing model and causes inconsistencies in 3D projection. Others~\cite{palandra2024gsedit} directly update Gaussians using Score Distillation Sampling (SDS) loss, but struggle to make significant modifications due to its implicit characteristic~\cite{chen2024mvdrag3d, chen2024gaussianeditor, guo2023stabledreamer, katzir2023noise}. We first remove priors and set the modification direction with the SLaMP-edited image, then refine for greater control beyond the original context. For masking strategies, most works utilize 2D masks for localized edits, projecting them onto Gaussians~\cite{chen2024gaussianeditor, wang2024gaussianeditor, wu2024gaussctrl}. However, noisy multi-view 2D masks introduce inconsistencies, affecting unintended regions or preventing proper transfer of 2D changes to 3D. Also, Gaussians at part boundaries can represent different parts depending on the view. However, assigning 3D Gaussian labels based on a 2D map overlooks this, resulting in inaccurate segmentation at part boundaries. To address this, our 3D-GALP selects anchors based on view-dependent label prediction consistency and enforces neighbor consistency in 3D, refining 3D masks to correct 2D imperfections. 

\subsection{Local editing of 3D representations} Most 3D editing methods discussed in Sec.~\ref{Sec_2.2} focus on instance level modifications or scene wide style changes. Some extend this to local edits, enabling precise adjustments to specific parts for finer control and prompt adaptability. A key challenge in local editing is effectively selecting specific areas. Some methods~\cite{cheng2023progressive3d, 10.1145/3664647.3681039} use bounding boxes from users or Large Language Models (LLMs) to make local changes, but these restrict selection to simple shapes, and their fixed nature prevents deformable edits. Another work~\cite{hyung2023local} utilizes a pre-trained 3D GAN~\cite{chan2022efficient} with CLIP~\cite{radford2021learning} to select local areas and generate changes in human and cat faces. While effective for some edits, it remains limited to specific targets and struggles with more drastic edits. Our model is the first to achieve part-level 3D editing for general objects in both reconstructed and generated Gaussians. By fully utilizing SD3 and Gaussian properties, RoMaP enables faithful and drastic 3D local edits.

\section{Method}
\label{sec:method}

 We propose RoMaP, a novel method for locally editing 3D Gaussians with text prompts, enabling targeted regional modifications. Existing approaches struggle with part edits because (1) projecting 2D segmentations to 3D is unreliable due to inconsistent part-aware models and ambiguous part boundaries, and (2) isolating specific parts is difficult due to entanglements in 2D diffusion models.


To address these challenges, RoMaP first performs explicit local 3D segmentation by adopting view-dependent segmentation labels and resolving 2D segmentation inconsistencies using 3D Geometry-Aware Label Prediction (3D-GALP), as discussed in Sec.~\ref{mask_gen}. To enable drastic part edits beyond pre-existing contexts, we introduce a new modification direction using regularized score distillation sampling, guided by regularizers: anchored $\mathcal{L}_1$ with Scheduled Latent Mixing and Part (SLaMP) editing, Gaussian prior removal and masking. This process is detailed in Sec.~\ref{coarse_2_fine}. The full pipeline of RoMaP is shown in Fig.~\ref{fig:pipeline}.

\subsection{Preliminary: 3D Gaussian Splatting}
\label{sec:preliminary} Gaussian Splatting~\cite{kerbl20233d} is a point-based method that represents a 3D scene using Gaussian properties. Let $\Omega$ be a set of Gaussians composing the scene, where each Gaussian $\Omega_i$ is defined as  
$\Omega_i = \{ \mathbf{p}_i, \mathbf{s}_i, \mathbf{q}_i, \alpha_i, \mathbf{c}_i \},$  
where $\mathbf{p}_i$, $\mathbf{s}_i$, $\mathbf{q}_i$, $\alpha_i$, and $\mathbf{c}_i$ represent the centroid, standard deviations, rotational quaternion, opacity, and spherical harmonics (SH) coefficients, respectively. The projected RGB color of Gaussians varies by viewpoint $\phi$ and is computed as ${\mathbf{c}}^{\phi} =SH(\mathbf{c}, \phi)$, where $SH(\mathbf{c}, \phi)$ evaluates the SH coefficients $\mathbf{c}$ at $\phi$. The rendered image ${\textbf{C}}^{\phi}$ for view $\phi$ is obtained by projecting $\Omega$ onto a 2D plane using the differentiable rasterization $\mathcal{D}$:
\vspace{-0.2cm}
\begin{center}
\begin{tikzpicture}
    \node (G) at (0, 0) {$\Omega=\{{\boldsymbol{p}, \boldsymbol{s}, \boldsymbol{q}, \alpha, \boldsymbol{c}}\}$};
    \node (S) at (3.2, 0) {$\mathbf{c}^{\phi} = SH(\boldsymbol{c}, \phi)$};
    \node (I) at (6.2, 0) {$\mathbf{C}^{\phi}$.};

    \draw[->] (S) -- node[above] {$\mathcal{D}$} (I);
    \draw[->] (1.1, 0.2) .. controls (1.1, 0.5) and (2, 0.5)
        .. node[below, xshift=0pt, yshift=2pt] {$\scriptstyle \phi$} (2, 0.2);
\end{tikzpicture}
\end{center}
\vspace{-0.4cm}

\subsection{Local 3D segmentation: 3D-GALP}
\label{mask_gen} This section describes the `Local 3D Segmentation' on the left side of the Fig.~\ref{fig:pipeline}. To localize changes in the target region, we create a 3D segmentation $\mathcal{M}_{3D}$ given $\Omega$. The goal is to predict which regions of $\mathcal{M}_{3D}$ correspond to each predefined part label $l_j$. This involves two steps: attention map extraction and 3D geometry-aware label prediction (3D-GALP). Given a segmentation prompt, we extract the attention map $\mathcal{A}(\mathbf{C}^{\phi})$ from a randomly rendered view $\mathbf{C}^{\phi}$ and treat it as a pseudo 2D segmentation map to guide 3D-GALP. More details on attention map extraction are in the supplementary material.

\paragraph{Attention-based pseudo segmentation for 3D Gaussians} 
In this step, we obtain the explicit 3D segmentation $\mathcal{M}_{3D}$ using 3D-GALP, guided by $\mathcal{A}(\mathbf{C}^{\phi})$. Once constructed, $\mathcal{M}_{3D}$ provides segmentation information for all Gaussians. To represent these labels, we introduce a new parameter $\mathbf{r}_{i}$ and incorporate it into the Gaussian representation: $\Omega_i = \{\mathbf{p}_i, \mathbf{s}_i, \mathbf{q}_i, \alpha_i, \mathbf{c}_{i}, \mathbf{r}_{i}\}$. Since a single Gaussian may correspond to different labels depending on the viewpoint, it exhibits mixed-label property. To model this view-dependent labeling, we represent each Gaussian's label as SH coefficients. We interpret $\mathbf{R^{\phi}}$, the 2D projection of Gaussians obtained via $\mathcal{\mathbf{r}}$ at view $\phi$, as a segmentation map:
\vspace{-0.4cm}
\[
\begin{tikzpicture}
    \node (G) at (0, 0) {$\Omega=\{{\boldsymbol{p}, \boldsymbol{s}, \boldsymbol{q}, \alpha, \boldsymbol{c}}, \textcolor{blue}{\boldsymbol{r}}\}$};
    \node (S) at (3, 0) {$\textcolor{blue}{\mathbf{r}}^{\phi} = SH(\textcolor{blue}{\boldsymbol{r}}, \phi)$};
    \node (I) at (5.3, 0) {$\boldsymbol{\textcolor{blue}{\mathbf{R}}^{\phi}}$.};
    \draw[->] (S) -- node[above] {$\mathcal{D}$} (I);
    \draw[->] (1.2, 0.2) .. controls (1.2, 0.5) and (2, 0.5) .. node[below, xshift=0pt, yshift=2pt] {$\scriptstyle \phi$} (2, 0.2);
\end{tikzpicture}
\]
\vspace{-0.6cm}


The learnable parameter $\mathcal{\mathbf{r}}$ is then optimized via \( \mathcal{L}_1(\mathcal{A}(\mathbf{C}^{\phi}), \mathbf{R}^{\phi})\) loss, encouraging the rendered map to align appropriately with the pseudo 2D attention map in the given view.
While this process aligns Gaussians with the attention map across multiple views, the alignment may remain imperfect. To further refine segmentation, we apply an anchor-based neighbor consistency loss, with anchors sampled by considering label softness. 

\begin{figure}[t]
    \centering
    \includegraphics[width=\linewidth]{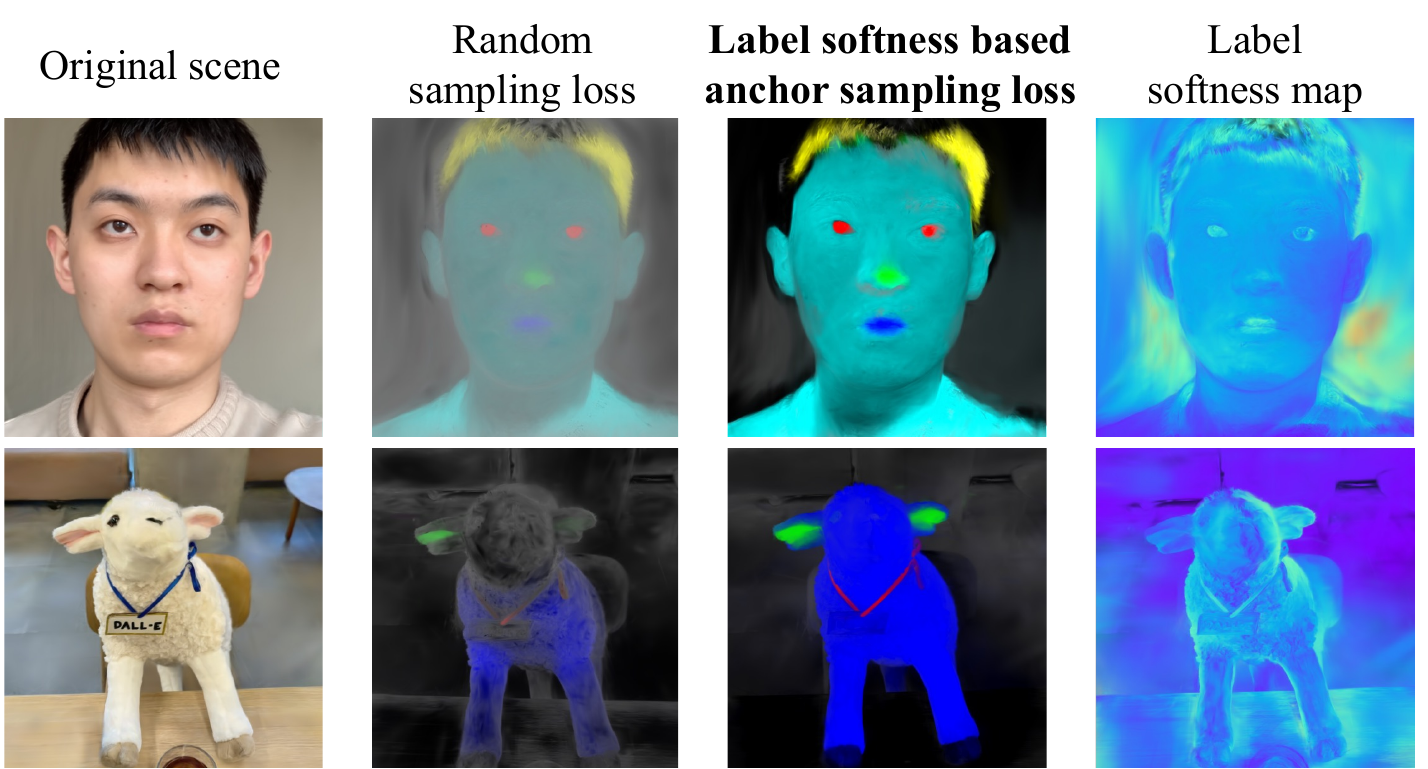}
     \vspace{-0.5cm}
    \caption{\textbf{Effectiveness of label softness-based anchor sampling.} By applying 3D loss with anchors sampled based on label softness, we observe that differentiation of boundaries between parts is much more precise compared to random sampling.}         
    \vspace{-0.4cm}
    \label{fig:3d-uncertainty}
\end{figure}

\paragraph{Label-softness based anchor sampling} Occlusions and view-dependent shape complexity can lead $\mathcal{A}(\mathbf{C}^{\phi})$ to produce incomplete segmentation maps (See Fig.~\ref{fig:pipeline}). To achieve complete and view-consistent 3D segmentation, we refine the segmentation by leveraging the view-dependent label softness of Gaussians. Here, $\mathbf{r}_i$ is treated as an SH color, and a Gaussian is considered to exhibit label softness if $\mathbf{r}_i^{\phi}$ varies with the viewpoint $\phi$. To quantify the label softness, we measure $v_i$, the variance of $\mathbf{r}^{\phi}$ across $\phi$. Then, we calculate the cosine similarity between \( \bar{\mathbf{r}}_i \), the mean color observed from all directions and \( \mathbf{l}_j \), the label assigned to each part. We then compute the entropy as follows:
\begin{equation}
\begin{aligned}
p_{ij} = \frac{e^{\frac{\bar{\mathbf{r}}_i \cdot \mathbf{\mathbf{l}}_j}{\|\bar{\mathbf{r}}_i\| \|\mathbf{\mathbf{l}}_j\|}}}{\sum_l e^{\frac{\bar{\mathbf{r}}_i \cdot \mathbf{l}_j}{\|\bar{\mathbf{r}}_i\| \|\mathbf{l}_j\|}}}, 
H_i = - \sum_j p_{ij} \log (p_{ij})
\end{aligned}
\label{eq:entropy}
\end{equation}
where \(p_{ij}\) denotes the probability  obtained from the cosine similarity between predicted label $\mathbf{r}_i$ and ground truth label $\mathbf{l}_j$, while \(H_i\) denotes the entropy of $p_{ij}$. 
We define the softness of the label of each Gaussian as the product of \( H_i \) and \( v_i \), given by \( S_i = H_i \cdot v_i \). As visualized in Fig.~\ref{fig:3d-uncertainty}, $S_i$ is high at part boundaries, where Gaussians inherently exhibit soft-label properties. This is due to the 2D part segmentation map classifying Gaussians noisly around these boundaries. All Gaussians are sorted based by $S_i$, then $K$ anchors are selected: the top \( \lfloor K/2 \rfloor \) from those with the highest softness values and the bottom \( \lfloor K/2 \rfloor \) from those with the lowest. This sampling method selects anchors from both Gaussians with high soft-label properties and those with consistent labels, enabling refinement of 3D segmentation while preserving locality and effectively handling part boundaries. Fig.~\ref{fig:3d-uncertainty} shows that part boundaries can be segmented precisely with label-softness based sampling compared to random sampling. 

\begin{figure*}[t]
    \centering
    \includegraphics[width=\textwidth]{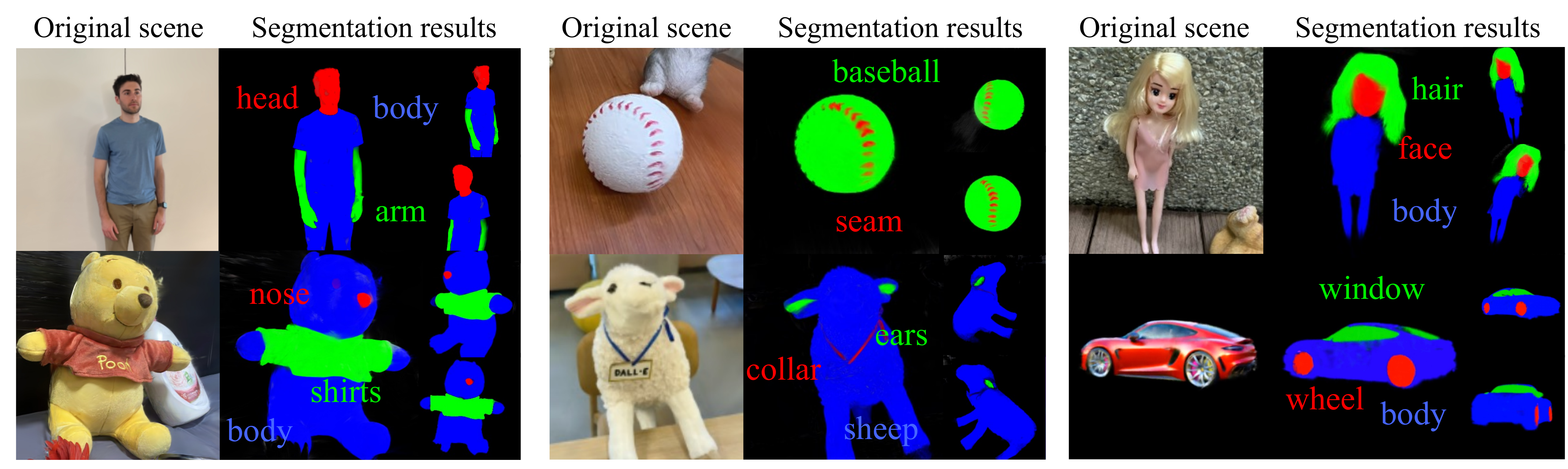}
    \vspace{-0.8cm}
    \caption{\textbf{3D Gaussian segmentation results of 3D-GALP.} With our 3D-GALP, 3D Gaussian segmentation accurately captures diverse object parts, addressing the limitations of 2D part segmentation and the inherent mixed nature of 3D Gaussian segmentation labels.}
     \vspace{-0.5cm}
    \label{fig:3d-seg}
\end{figure*}

\paragraph{Anchor-based neighboring loss}
Given the selected anchors $A$, we enforce neighbor consistency by incorporating segmentation information from nearby Gaussians. For each anchor $\Omega_i \in A$, we find its \( K \) nearest neighbors, where \( \mathcal{N}_K(i) \) denotes the top-\( k \) nearest neighbors of the \( i \)-th anchor. We then compute the $\mathcal{L}_1$ between the segmentation label \( \mathbf{r}_j \) of neighboring points and the \( \mathbf{r}_i \) of the anchor point:
\begin{equation}
\begin{aligned}
\mathcal{L}_{\text{GALP}} = \sum_{i \in A} \left[ \frac{1}{K} \sum_{k \in \mathcal{N}_K(i)} \| \mathbf{r}_{i} - \mathbf{r}_{k} \|_1 \right].
\end{aligned}
\label{eq:neighbor}
\end{equation}
As shown in Fig.~\ref{fig:3d-seg}, 3D-GALP effectively can segment various parts of diverse objects. Additional 3D segmentation results in various scenes are provided in the supplementary.

\subsection{Local 3D Editing: Regularized score distillation sampling}
\label{coarse_2_fine}

\begin{figure}
    \centering
     \includegraphics[width=1 \columnwidth]{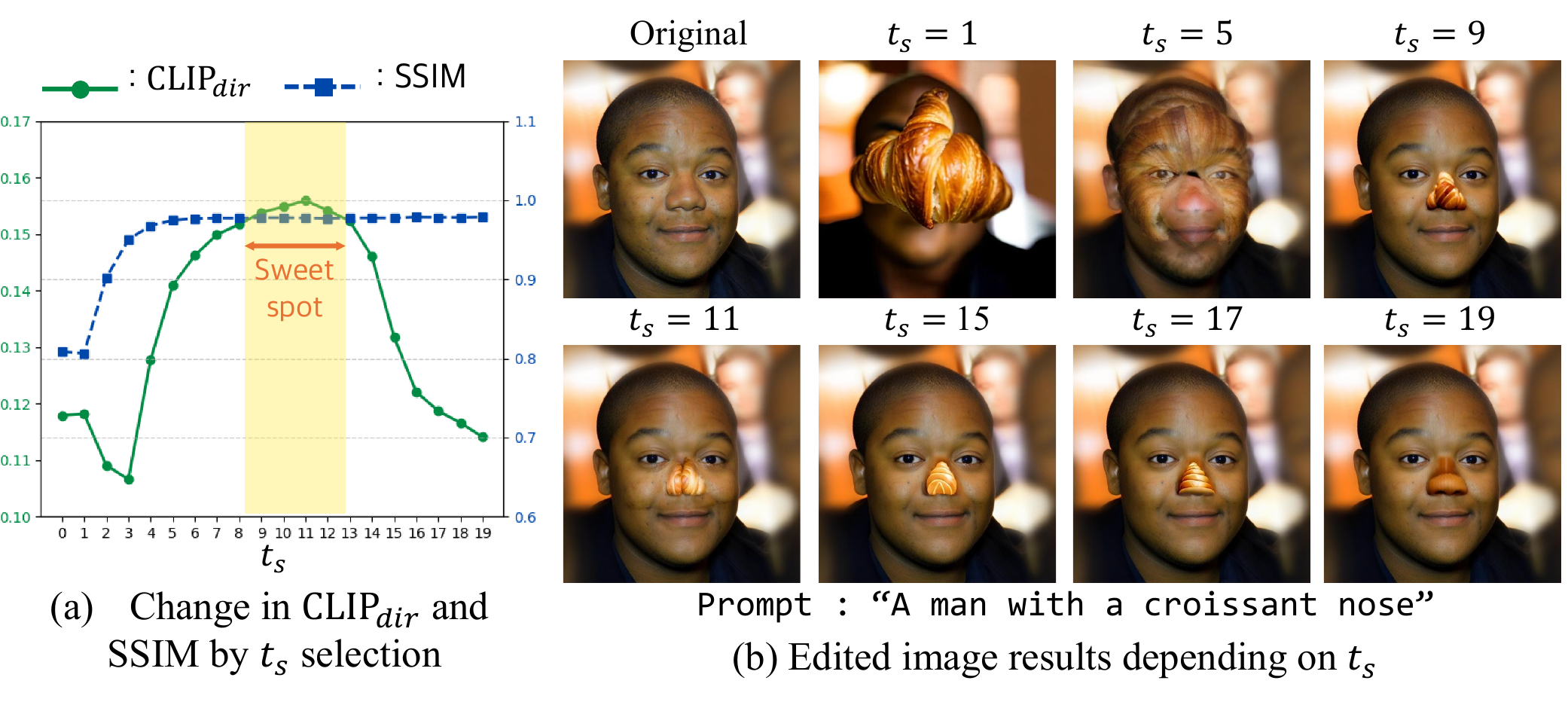}
          \vspace{-0.8cm}
    \caption{\textbf{Experiments on the effect of $t_s$.} $t_s$ controls the extent of deviation from the original. We set $t_s$ to induce drastic changes in the target region while preserving the surrounding identity.}
          \vspace{-0.5cm}
    \label{fig:latentmixing}
\end{figure}

\paragraph{Regularized score distillation sampling} We can now explicitly select Gaussian regions for editing using $\mathcal{M}_{3D}$. Since the SDS loss primarily serves as an implicit objective but has limited direct impact on 3D Gaussians~\cite{chen2024mvdrag3d, chen2024gaussianeditor, guo2023stabledreamer, katzir2023noise}, we enable more effective modifications by introducing a regularized SDS loss. This loss combines two regularizers: Gaussian prior removal and masking, and an anchored-based $\mathcal{L}_1$ loss using SLaMP edited image. The regularized SDS loss is defined as: 
\vspace{-0.1cm}
\begin{equation}
\mathcal{L}_{R\text{-}SDS} = \lambda_1\hat{\mathcal{L}}_{SDS}(c^{\phi}_{pr}, \textbf{p}_{\text{edit}}) + \lambda_2 \hat{\mathcal{L}_1}(c^{\phi}_{pr}, \text{SLaMP}(c^{\phi}_{pr})).
\label{eq:latentmix}
\end{equation}
Here, $\lambda_1$ and $\lambda_2$ are hyperparameters that balance the contribution of $\hat{\mathcal{L}}_{SDS}$ and $\hat{\mathcal{L}}_{1}$ during training. 
$\hat{\mathcal{L}}$ denotes a masked loss leveraging $\mathcal{M}_{3D}$ and $\mathcal{M}_{2D}$ to restrict changes only to intended regions.
$c^{\phi}_{pr}$ refers to the 2D projection of prior-removed Gaussians in view $\phi$ and \text{SLaMP} refers to our 2D part editing method that enables clear directional change that SDS loss cannot achieve. These two components will be discussed in following section.

\vspace{-0.3cm}
\paragraph{Regularizer 1: Gaussian prior removal and masking} 
Due to the inherent ambiguity of SDS loss and the localized nature of Gaussians, applying SDS alone limits modification extent~\cite{guo2023stabledreamer, chen2024mvdrag3d}. To address this, we introduce an $\mathcal{L}_1$ loss on explicitly edited images to provide more targeted and controllable guidance. However, directly combining $\mathcal{L}_1$ with $\mathcal{L}_{SDS}$ often induces overly broad changes, since $\mathcal{L}_{SDS}$ operates in all directions and biases the optimization toward preserving strong appearance priors. To mitigate this, we perform Gaussian prior removal by replacing dominant color priors with neutral colors (\textit{e.g.}, white or gray), producing $c^{\phi}_{pr}$ to discourage fixation on original appearances. Additionally, we explicitly prevent gradient updates to Gaussians on $\mathcal{M}_{3D}$, avoiding unintended changes and ensuring that edits are confined to the target regions.

\paragraph{Regularizer 2: Anchored $\mathcal{L}_1$ with SLaMP edited image} To generate an anchor image for the $\mathcal{L}_1$ loss, we propose SLaMP editing, a part level editing strategy that balances localized modification with global identity preservation. A key aspect of SLaMP is the scheduled blending of latents over time, enabling fine-grained control over the influence of the original image. Effective part-level editing requires isolating the target region while guiding it toward the desired change without compromising global identity. SLaMP achieves this by scheduling a sharp transition in the blending ratio between the target latent $\mathbf{z}_{t}$ and the original latent $\mathbf{z}_{t, \text{orig}}$. The resulting latent $\mathbf{z}_{t+1}$ is expressed as follows:
\begin{equation}
\mathbf{z}_{t+1} = \mathbf{z}_{t} (1 - \mathcal{F}_t  \cdot (1 - \mathcal{M}_{2D})) 
+ \mathbf{z}_{t, \mathrm{orig}}  \mathcal{F}_t \cdot (1 - \mathcal{M}_{2D}).
\label{eq:latentmix}
\end{equation}
Here, \( \mathcal{F}_t \) is a time-dependent blending coefficient. We begin with a low \( \mathcal{F}_t\) to generate new context without strong influence from the original image. At timestep \( t_s \), we increase \( \mathcal{F}_t\) sharply to preserve the alignment with original. As shown in Fig.~\ref{fig:latentmixing}, setting \( t_s \) too low disrupts the original image context, while setting it too high hinders new content generation. To balance preservation and editing, we set $t_s$ to where SSIM is stable while CLIP$_{dir}$ remains high. More details are in the supplementary.

\begin{figure*}[!ht]
    \centering
    \vspace{-0.8cm}
     \includegraphics[width=1.0 \textwidth]{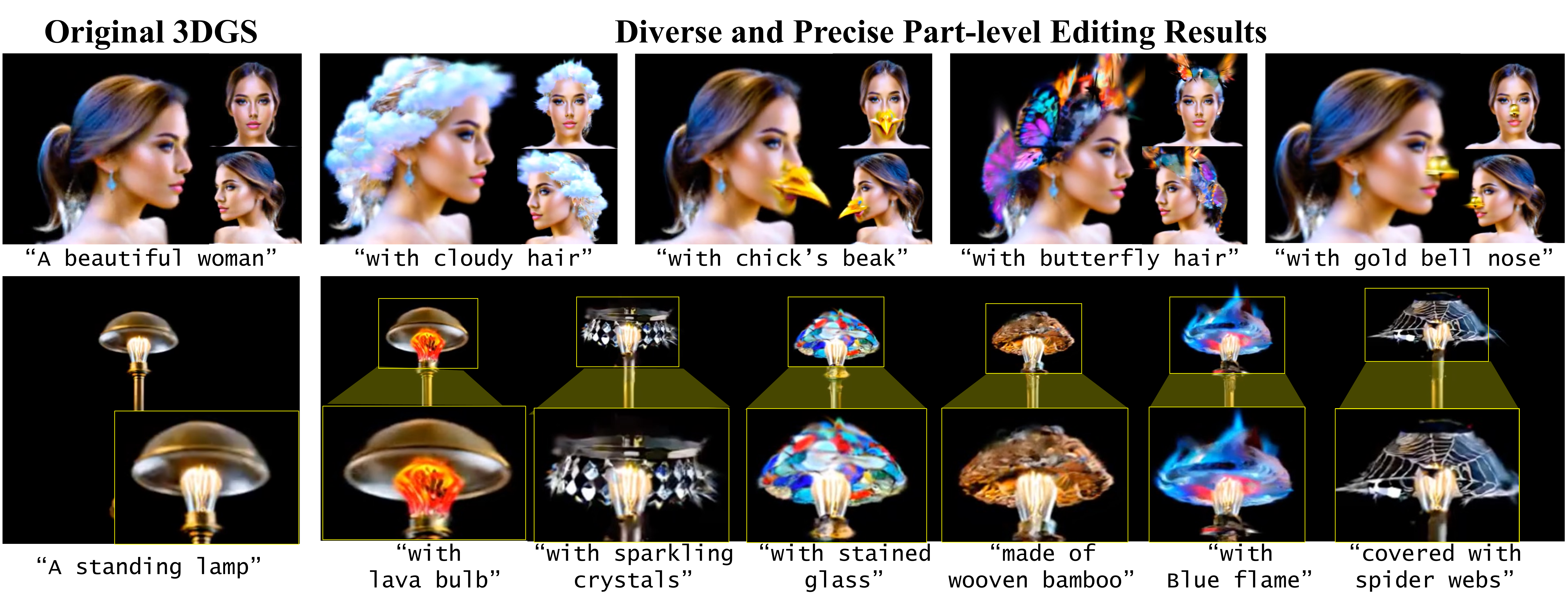}
      \vspace{-0.8cm}
    \caption{\textbf{Enhanced controllability in 3D asset generation with RoMaP.} Our approach enables precise manipulation of specific 3D parts. As shown above, RoMaP provides diverse control over multiple narrow regions within a single 3D object, allowing deformations in targeted areas like a `duck's beak' or `jellyfish hair' and facilitating various modifications in targeted area such as a lamp's lampshade.}
   \vspace{-0.5cm}
    \label{fig:controllability}
\end{figure*}

\section{Experiments}
\subsection{Experimental setting} 
\textbf{Dataset and evaluation metrics} To evaluate editing performance on reconstructed Gaussians, we use scenes from IN2N~\cite{haque2023instruct} and NeRF-Art~\cite{haque2023instruct}, testing 75 editing prompts targeting different parts and changes in each scene. For evaluation metrics, we used two CLIP-based metrics, CLIP Similarity~\cite{radford2021learning} and CLIP$_{dir}$ Similarity~\cite{gal2022stylegan}, to measure the overall fidelity between the input text and the edited scene. Furthermore, we used BLIP-VQA~\cite{huang2023t2i} and TIFA~\cite{hu2023tifa} to assess how well edits align with specific text prompt components via visual question answering.

\noindent \textbf{Baselines} We compared RoMaP with three state-of-the-art 3D Gaussian editing methods (DGE~\cite{chen2024dge}, GaussianEditor~\cite{chen2024gaussianeditor}, and GaussCtrl~\cite{wu2024gaussctrl}) and three NeRF editing methods (Instruct-Nerf2Nerf (IN2N)~\cite{haque2023instruct}, ViCA-NeRF (ViCA)~\cite{dong2023vica}, and Posterior Distillation Sampling (PDS)~\cite{Koo:2024PDS}). All baselines perform 2D edits before lifting them to 3D. \cite{chen2024gaussianeditor, chen2024dge, dong2023vica} employ InstructPix2Pix~\cite{brooks2023instructpix2pix}, while~\cite{wu2024gaussctrl} utilizes ControlNet~\cite{zhang2023adding}, and~\cite{Koo:2024PDS} applies posterior distillation sampling. In a user study, we compared RoMaP against generation models~\cite{yi2023gaussiandreamer,chen2024text,yang2024text}, assessing how editing improves controllability in generating previously difficult samples. 

\subsection{Experimental results} 
\begin{table}[t]
\resizebox{\linewidth}{!}{ 
\begin{tabular}{l|cccc}
\toprule
\multicolumn{1}{c|}{\textbf{Editing Methods}} & \multicolumn{4}{c}{\textbf{Metrics}} \\ \cmidrule(lr){2-5}  
 & CLIP $\uparrow$  & CLIP$_{dir}$ $\uparrow$ & B-VQA $\uparrow$ & TIFA $\uparrow$ \\ \midrule
\rowcolor{gray!15} \multicolumn{5}{l}{\textbf{NeRF baselines}} \\

IN2N~\cite{haque2023instruct}             & 0.248 & 0.072     & 0.142 & 0.634  \\
ViCA~\cite{dong2023vica} & 0.223 & 0.048    & 0.241 & 0.427  \\
PDS~\cite{Koo:2024PDS}               & 0.167 & -0.005     & 0.237 & 0.212  \\ \midrule
\rowcolor{gray!15} \multicolumn{5}{l}{\textbf{Gaussian Splatting baselines}} \\ 
GaussCtrl~\cite{wu2024gaussctrl}             & 0.182 & 0.044     & 0.190 & 0.432  \\
GaussianEditor~\cite{chen2024gaussianeditor} & 0.179 & 0.087     & 0.370 & 0.571 \\
DGE~\cite{chen2024dge}                       & 0.201 & 0.095     & 0.497 & 0.565  \\
\rowcolor{yellow!25} \textbf{RoMaP (Ours)}  & \textbf{0.277} & \textbf{0.205}     & \textbf{0.723} & \textbf{0.674} \\ 
\bottomrule
\end{tabular}
}
\caption{\textbf{Quantitative comparison with 3D editing methods.} Our method outperforms various baselines in multiple metrics.}
\vspace{-0.4cm}
\label{tab:quant_results}
\end{table}

\textbf{Quantitative comparisons} Tab.~\ref{tab:quant_results} presents a quantitative comparison of RoMaP against 3DGS and NeRF editing models, where it outperforms all baselines across metrics. As shown in Tab.~\ref{tab:user_study}, user study further validates RoMaP's superior performance. Statistical significance of user study is confirmed by Friedman and pairwise Wilcoxon tests.

\begin{table}[!ht]
\centering
\resizebox{\linewidth}{!}{\begin{tabular}{l|c|l|c}
\toprule
Editing Method & User Study $\uparrow$ & Generation Method & User Study $\uparrow$ \\ \midrule
GaussCtrl~\cite{wu2024gaussctrl} & 0.201  & GSGEN~\cite{chen2024text} & 0.203 \\
GaussianEditor~\cite{chen2024gaussianeditor} & 0.201  & GaussianDreamer~\cite{yi2023gaussiandreamer} & 0.198 \\
DGE~\cite{chen2024dge}  & 0.224  & RFDS~\cite{yang2024text} & 0.234 \\ 
\rowcolor{yellow!25}
\textbf{RoMaP (Ours)}  & \textbf{0.372} & \textbf{RoMaP (Ours)}  & \textbf{0.365} \\
\bottomrule
\end{tabular}}
\caption{\textbf{User study results.} Quantitative comparison of user study results for editing and generation methods.}
\vspace{-0.4cm}
\label{tab:user_study}
\end{table}

\begin{figure*}
    \centering
     \includegraphics[width=0.76 \textwidth]{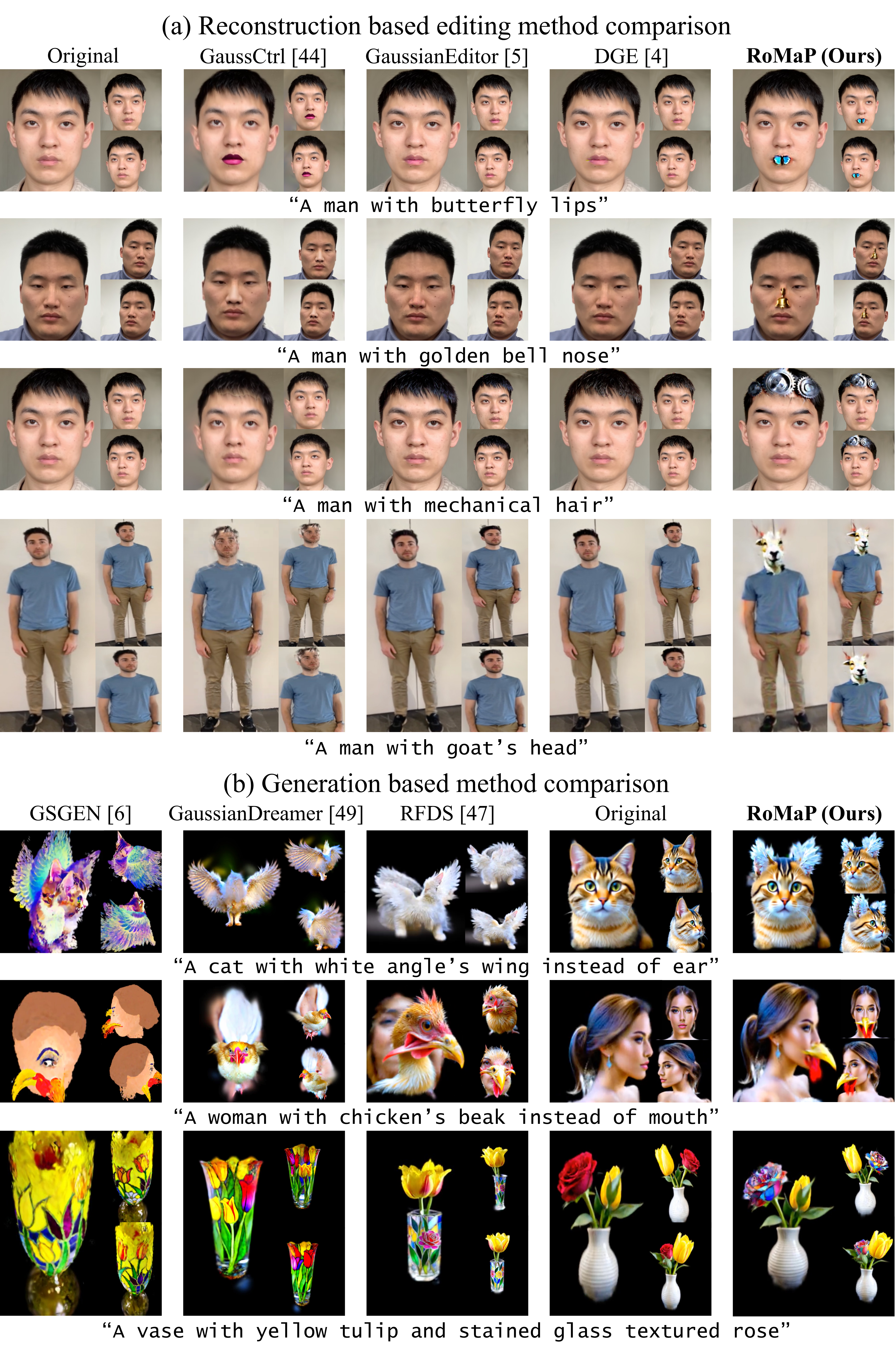}
    \caption{\textbf{Comparison results} The results of comparing our methodology with various reconstruction-based 3D editing methods and text-to-3D generation approaches are presented. In the reconstructed scene, our method enables drastic changes in very narrow regions, breaking the existing priors that other approaches have been unable to overcome. This allows for diverse transformations, such as replacing a human face with a goat's face or substituting hair with butterflies. In the text-to-3D generation scenario, our approach achieves success in examples where naive text prompts alone fail, demonstrating its ability to generate a wider range of 3D assets.}
    \label{fig:comparison}
\end{figure*}

\noindent \textbf{Qualitative comparisons} Fig.~\ref{fig:comparison} shows qualitative results comparing RoMaP with 3DGS generation and editing methods. Ours enables drastic local changes, such as butterfly lips and goat's head, while others fail. Its enhanced controllability also enables text-aligned generation that other models struggle with. As shown in Fig.~\ref{fig:controllability}, RoMaP enables diverse 3D creations, such as a lamp with different bulbs and lampshades, simplifying 3D asset customization.

\begin{table}[h]
\centering
\resizebox{1\linewidth}{!}{ 
\begin{tabular}{l|c|c|c|c}
\toprule
\textbf{Metrics} & \textbf{Baseline} & \textbf{+ Mask} & \textbf{+ Mask \& \(\hat{\mathcal{L}_1}\)} & \cellcolor{yellow!20}\textbf{Full (Ours)} \\
\midrule
CLIP $\uparrow$       & 0.218 & 0.228 & 0.267 & \cellcolor{yellow!20}\textbf{0.277} \\
CLIP$_{dir} \uparrow$ & 0.060 & 0.162 & 0.205 & \cellcolor{yellow!20}\textbf{0.205} \\
\bottomrule
\end{tabular}
}
\caption{\textbf{Ablation study results} The ablation study shows results from sequentially adding key methods. 
}
\vspace{-0.4cm}
\label{tab:abl}
\end{table}

\subsection{Ablation study} Tab.~\ref{tab:abl} presents an ablation study validating each step of RoMaP. In Tab.~\ref{tab:abl}, `Mask' refers to results using masks (\( \mathcal{M}_{2D} \) \& \( \mathcal{M}_{3D} \)) generated from 3D-GALP. `\(\hat{\mathcal{L}_1}\)' is the result of a regularized SDS loss, by only employing the second term. The `Full' represent our full regularized SDS loss, enabling modification in the desired direction. This confirms the necessity of all steps in RoMaP. 

\section{Conclusion}
In this work, we introduce RoMaP, a novel approach for local 3D Gaussian editing that enables precise and consistent part-level edits. To localize part accurately, we employ robust segmentation with geometry-aware label prediction, utilizing the soft-label properties of Gaussians. We also propose the regularized SDS loss using scheduled latent mixing and Gaussian prior removal, enabling drastic part-level edits while preserving remaining areas. Experimental results demonstrate RoMaP's significant improvements in 3D Gaussian editing quality across various scenes even in challenging scenarios. 

\section*{Acknowledgements}
This work was supported in part by Institute of Information \& communications Technology Planning \& Evaluation (IITP) grant funded by the Korea government(MSIT) [NO.RS-2021-II211343, Artificial Intelligence Graduate School Program (Seoul National University)], the National Research Foundation of Korea(NRF) grant funded by the Korea government(MSIT) (No. RS-2025-02263628) and AI-Bio Research Grant through Seoul National University. Also, the authors acknowledged the financial support from the BK21 FOUR program of the Education and Research Program for Future ICT Pioneers, Seoul National University.

\clearpage
\setcounter{section}{0}
\setcounter{figure}{0}
\setcounter{table}{0}
\setcounter{equation}{0}
\renewcommand{\thesection}{S.\arabic{section}}
\renewcommand{\thefigure}{S.\arabic{figure}}
\renewcommand{\thetable}{S.\arabic{table}}
\renewcommand{\theequation}{S.\arabic{equation}}

\twocolumn[%
\vspace*{2em}
\begin{center}
\begin{minipage}{0.94\textwidth}
  \centering
    {\Large \textbf{Supplementary Material for}}%
    \vspace{0.8em}
    
    {\Large \textbf{Robust 3D-Masked Part-level Editing in 3D Gaussian Splatting}}%
    \vspace{0.3em}
    
    {\Large \textbf{with Regularized Score Distillation Sampling}}
\end{minipage}
\end{center}
\vspace{2em}
]

\begin{table*}[tbp]
\centering
\small
\resizebox{\textwidth}{!}{
\begin{tabular}{c|c|c|c|c|c|c|c|c|c|c|c|c}
\toprule
  \multirow{3}{*}{method}    & \multicolumn{12}{c}{part}                                                                         \\ \cmidrule{2-13} 
 & \multicolumn{2}{c|}{eye} & \multicolumn{2}{c|}{nose} & \multicolumn{2}{c|}{mouth} & \multicolumn{2}{c|}{hair} & \multicolumn{2}{c|}{hard} & \multicolumn{2}{c}{avg} \\ \cmidrule{2-13} 
                        & $\scalebox{0.8}{CLIP}$     & $\scalebox{0.8}{CLIP$_{dir}$}$    & $\scalebox{0.8}{CLIP}$     & $\scalebox{0.8}{CLIP$_{dir}$}$    & $\scalebox{0.8}{CLIP}$     & $\scalebox{0.8}{CLIP$_{dir}$}$     & 
                        
                        & $\scalebox{0.8}{CLIP}$    & $\scalebox{0.8}{CLIP$_{dir}$}$    & $\scalebox{0.8}{CLIP}$   & $\scalebox{0.8}{CLIP$_{dir}$}$   \\ \midrule \midrule
GaussCtrl~\cite{wu2024gaussctrl}               & 0.191    &   0.042        & 0.183     &   0.035        & 0.173    &   0.056        & 0.195     &    0.060      & 0.168     &     0.026         & 0.182   &      0.044   \\ \midrule
GaussianEditor~\cite{chen2024gaussianeditor}          & 0.190    &    0.068    & 0.130     &     0.057    & 0.140     &      0.086    & 0.232     &    0.144    & 0.202     &   0.083     & 0.179    &  0.087         \\ \midrule
DGE~\cite{chen2024dge}                     & 0.193    &    0.076     & 0.190     &    0.058    & 0.182     &   0.070     & 0.232    &    0.161    &   0.211  &     0.110   & 0.201   &      0.095    \\ \midrule
\rowcolor{yellow!25} \textbf{RoMaP(ours) }           & \textbf{0.246}    &    \textbf{0.150}   & \textbf{0.263}     &  \textbf{ 0.210}     & \textbf{0.311}    &    \textbf{0.265}       & \textbf{0.277}    &  \textbf{ 0.211}    & \textbf{0.291}     &        \textbf{ 0.188}   & \textbf{0.277}    &  \textbf{ 0.205}    \\ \bottomrule
\end{tabular}}
\caption{\textbf{Comparison with GS editing methods.} CLIP score and CLIP directional score value for each method and part.}
\label{table:suptable1}
\end{table*}

\begin{table*}[tbp]
\small
\resizebox{\textwidth}{!}{
\begin{tabular}{c|c|c|c|c|c|c|c|c|c|c|c|c}
\toprule
  \multirow{3}{*}{method}    & \multicolumn{12}{c}{part}                                                                                                                                      \\ \cmidrule{2-13} 
 & \multicolumn{2}{c|}{eye} & \multicolumn{2}{c|}{nose} & \multicolumn{2}{c|}{mouth} & \multicolumn{2}{c|}{hair} & \multicolumn{2}{c|}{hard} & \multicolumn{2}{c}{avg} \\ \cmidrule{2-13} 
                        & $\scalebox{1}{B-VQA}$     & $\scalebox{1}{TIFA}$    & $\scalebox{1}{B-VQA}$     & $\scalebox{1}{TIFA}$    & $\scalebox{1}{B-VQA}$     & $\scalebox{1}{TIFA}$     & $\scalebox{1}{B-VQA}$    & $\scalebox{1}{TIFA}$   & $\scalebox{1}{B-VQA}$    & $\scalebox{1}{TIFA}$    & $\scalebox{1}{B-VQA}$   & $\scalebox{1}{TIFA}$   \\ \midrule \midrule
GaussCtrl~\cite{wu2024gaussctrl}               &0.194    & 0.422             &0.195      &0.561              &0.223      &0.389               &0.239      &0.494              &0.098      &0.292              &0.190     &0.432              \\ \midrule
GaussianEditor~\cite{chen2024gaussianeditor}          &0.361     &0.561              &0.301     &0.633              &0.448     &0.572               &0.593      &0.722             &0.148     &0.368              &0.370     &0.571              \\ \midrule
DGE~\cite{chen2024dge}                     &0.517     &0.539             &0.427      &0.717             &0.512      &0.5              &0.774     &0.683              &0.255      &0.388              &0.497     &0.565             \\ \midrule
\rowcolor{yellow!25} \textbf{RoMaP(ours) }           &\textbf{0.700}     &\textbf{0.667}             &\textbf{0.797}     &\textbf{0.733}              &\textbf{0.935}     &\textbf{0.711}              &\textbf{0.796}      &\textbf{0.717}              &\textbf{0.399}     &\textbf{0.543}              &\textbf{0.723}     &\textbf{0.674}             \\ \bottomrule
\end{tabular}
}
\caption{\textbf{Comparison with GS editing methods.} BLIP-VQA score and TIFA score value for each method and part.}
\label{table:suptable2}
\end{table*}

\begin{table*}[!h]
\small
\resizebox{\textwidth}{!}{
\begin{tabular}{c|c|c|c|c|c|c|c|c|c|c|c|c}
\toprule
  \multirow{3}{*}{method}    & \multicolumn{12}{c}{part}                                                                                                                                      \\ \cmidrule{2-13} 
 & \multicolumn{2}{c|}{eye} & \multicolumn{2}{c|}{nose} & \multicolumn{2}{c|}{mouth} & \multicolumn{2}{c|}{hair} & \multicolumn{2}{c|}{hard} & \multicolumn{2}{c}{avg} \\ \cmidrule{2-13} 
                        & $\scalebox{0.8}{CLIP}$     & $\scalebox{0.8}{CLIP$_{dir}$}$    & $\scalebox{0.8}{CLIP}$     & $\scalebox{0.8}{CLIP$_{dir}$}$    & $\scalebox{0.8}{CLIP}$     & $\scalebox{0.8}{CLIP$_{dir}$}$     & $\scalebox{0.8}{CLIP}$    & $\scalebox{0.8}{CLIP$_{dir}$}$   & $\scalebox{0.8}{CLIP}$    & $\scalebox{0.8}{CLIP$_{dir}$}$    & $\scalebox{0.8}{CLIP}$   & $\scalebox{0.8}{CLIP$_{dir}$}$   \\ \midrule \midrule
iN2N~\cite{haque2023instruct}               & \textbf{0.247} &   0.067       & 0.257     &   0.071       & 0.258    &   0.084        & 0.253     &    0.079      & 0.227     &     0.060         & 0.248     &  0.072     \\ \midrule
VICA~\cite{dong2023vica}          & 0.224    &    0.050    & 0.225     &   0.040    & 0.219     &      0.052    & 0.229     &    0.049    & 0.217     &   0.051     &   0.223   &  0.048         \\ \midrule
PDS~\cite{Koo:2024PDS}                     & 0.162    &    -0.033     & 0.171     &    0.014    & 0.177     &   0.007     & 0.176    &    0.008    &   0.152  &     -0.020   &  0.167  &  -0.005       \\ \midrule
\rowcolor{yellow!25} \textbf{RoMaP(ours) }           & 0.246    &    \textbf{0.150}   & \textbf{0.263}     &  \textbf{ 0.210}     & \textbf{0.311}    &    \textbf{0.265}       & \textbf{0.277}    &  \textbf{ 0.211}    & \textbf{0.291}     &        \textbf{ 0.188}   & \textbf{0.277}    &  \textbf{ 0.205}    \\ \bottomrule
\end{tabular}}
\caption{\textbf{Comparison with NeRF editing methods.} CLIP score and CLIP directional score value for each method and part.}
\label{table:suptable3}
\end{table*}

\begin{table*}[!h]
\small
\resizebox{\textwidth}{!}{
\begin{tabular}{c|c|c|c|c|c|c|c|c|c|c|c|c}
\toprule
  \multirow{3}{*}{method}    & \multicolumn{12}{c}{part}                                                                                                                                      \\ \cmidrule{2-13} 
 & \multicolumn{2}{c|}{eye} & \multicolumn{2}{c|}{nose} & \multicolumn{2}{c|}{mouth} & \multicolumn{2}{c|}{hair} & \multicolumn{2}{c|}{hard} & \multicolumn{2}{c}{avg} \\ \cmidrule{2-13} 
                        & $\scalebox{1}{B-VQA}$     & $\scalebox{1}{TIFA}$    & $\scalebox{1}{B-VQA}$     & $\scalebox{1}{TIFA}$    & $\scalebox{1}{B-VQA}$     & $\scalebox{1}{TIFA}$     & $\scalebox{1}{B-VQA}$    & $\scalebox{1}{TIFA}$   & $\scalebox{1}{B-VQA}$    & $\scalebox{1}{TIFA}$    & $\scalebox{1}{B-VQA}$   & $\scalebox{1}{TIFA}$   \\ \midrule \midrule
iN2N~\cite{haque2023instruct}               &0.168    & 0.589            &0.168      &0.489              &0.163      &0.471               &0.139      &0.671              &0.072      &\textbf{0.623}              & 0.142     &0.565         \\ \midrule
VICA~\cite{dong2023vica}          &0.277     &0.436              &0.204     &0.507              &0.292     &0.387               &0.228      &0.396             &0.205    &0.41              &0.241    &0.427            \\ \midrule
PDS~\cite{Koo:2024PDS}                     &0.267     &0.2             &0.287      &0.173             &0.264      &0.147              &0.333     &0.160              &0.034      &0.380              &0.237    &0.212             \\ \midrule
\rowcolor{yellow!25} \textbf{RoMaP(ours) }           &\textbf{0.700}     &\textbf{0.667}             &\textbf{0.797}     &\textbf{0.733}              &\textbf{0.935}     &\textbf{0.711}              &\textbf{0.796}      &\textbf{0.717}              &\textbf{0.399}     &0.543
&\textbf{0.723}     &\textbf{0.674}             \\ \bottomrule
\end{tabular}
}
\caption{\textbf{Comparison with GS editing methods.} BLIP-VQA score and TIFA score value for each method and part.}
\label{table:suptable4}
\end{table*}

\begin{table}[ht!]
\begin{tabular}{l|ccc}

\toprule
        Method       & Alignment & Fidelity & \multicolumn{1}{l}{Accuracy} \\ \midrule \midrule
GaussCtrl~\cite{wu2024gaussctrl}      & 19.70\%   & 19.98\%  & 20.6\%                              \\
GaussianEditor~\cite{chen2024gaussianeditor} & 19.61\%   & 19.98\%  & 20.72\%                             \\
DGE~\cite{chen2024dge}            & 23.18\%   & 23.62\%  & 20.24\%                             \\
\rowcolor{yellow!25} \textbf{RoMaP (Ours)}  & \textbf{36.73\%}   & \textbf{36.31\% } & \textbf{38.43\% }  \\ \bottomrule
\end{tabular}
\caption{\textbf{User study results on comparison with 3D Gaussian editing models.}}
\label{tab:recon_userstudy}
\end{table}

\begin{table}[ht!]
\centering
\begin{tabular}{l|cc}
\toprule
      Method          & Alignment & Fidelity \\ \midrule \midrule
GSGEN~\cite{chen2024text}           & 20.48\%   & 20.09\%   \\
GaussianDreamer~\cite{yi2023gaussiandreamer} & 19.61\%   & 19.98\%  \\
RFDS~\cite{yang2024text}            & 23.18\%    & 23.62\%   \\
\rowcolor{yellow!25}\textbf{RoMaP (Ours) }  & \textbf{36.73\%}   & \textbf{36.31\%} \\ \bottomrule
\end{tabular}
\caption{\textbf{User study results on comparison with 3D Gaussian generation models.}}
\label{tab:gen_userstudy}
\end{table}

\section{Additional details on quantitative results}
\label{sec:Additional_Exp}
\subsection{Experimental setting}
\subsubsection{Comparison with 3D Gaussian editing models}
\label{sec:quanti_setting}
We collected human face scenes from the IN2N~\cite{haque2023instruct} and Nerf-Art~\cite{wang2023nerf} datasets. For each facial part: `eyes', `nose', `mouth' and `hair', we applied five editing prompts: `silver-textured', `gold-textured', `diamond', `green', `pink' to evaluate editing success. Additionally, we designed five prompts requiring drastic changes: `delicious croissant nose', `hair made of metallic gears, steampunk style', `hair on fire, red and blue flame', `hair covered with beautiful butterfly', `left blue and right green eye', and categorized them as `hard' to assess extreme editing performance. For models incorporating InstructPix2Pix~\cite{brooks2023instructpix2pix} in their pipelines, we adapted the prompts to the format: ``Turn ... into ...''. 

\subsubsection{Comparison with 3D Gaussian generation models}
To prove that our local 3D editing method enhances controllability in 3D content generation, we designed prompts for samples that were challenging for previous 3D generation methods to create. The prompts included: `A beautiful woman with a cheek's beak', `A woman with cloudy hair', `A beautiful woman with butterfly hair', `A snail with skyscapes inside its shell', and `A vase with a yellow tulip and a stained glass-textured rose'. We tasked 3D generation models with directly generating 3D content from these prompts. In our approach, we first generated the base objects, such as `A snail', then applied these prompts as editing instructions to assess whether our method could successfully produce the desired samples. 

\paragraph{User Study} 
We conducted a user study across three categories: (1) Alignment - Is the 3D Gaussian edited to match the text? (2) Fidelity - Does the image look visually appleaing? (3)  Accuracy - Were only the specified parts edited correctly?. Users were asked to score a 4-point scale, and we averaged it for mean opinion score (MOS). For reconstructed scene, participants evaluated all three criteria, collecting 4,680 responses from 260 respondents. For generated 3D, evaluations were based on alignment and fidelity, yielding 2,600 responses from 260 respondents. 

\begin{figure*}
    \centering
     \includegraphics[width=0.8 \textwidth]{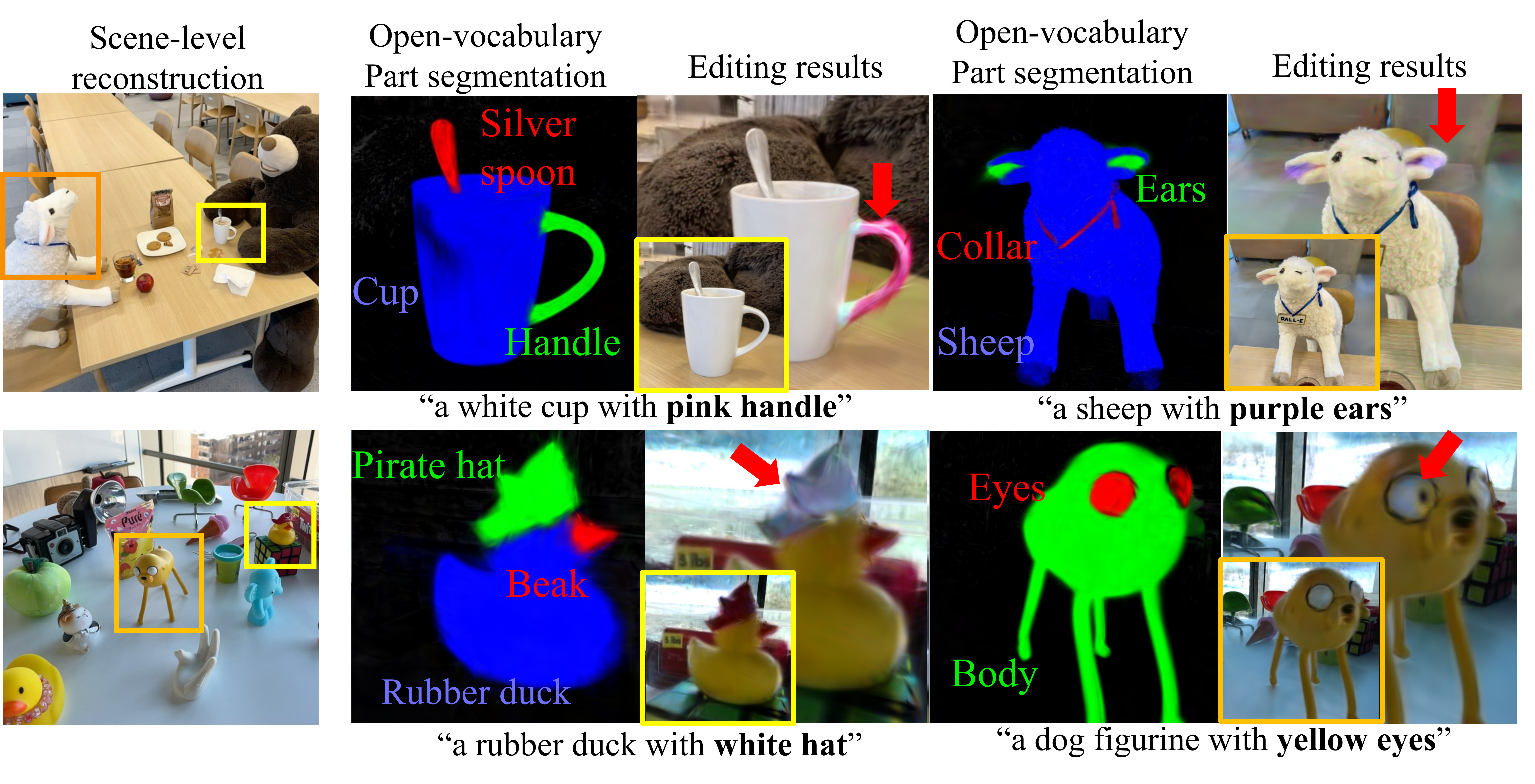}
    \caption{\textbf{3DGS part editing results in complex 3DGS scenes.} We performed RoMaP editing on complex 3DGS scenes from the LERF dataset. As shown above, our RoMaP achieved precise open-vocabulary part segmentation for parts of varying sizes, such as the collar, eyes, body, and rubber duck. Additionally, we achieved accurate part editing based on prompts like `a sheep with purple ears' and `a rubber duck with a white hat'. }
    \label{fig:comp_editing1}
\end{figure*}

\begin{figure*}
    \centering
    \includegraphics[width=0.8\textwidth]{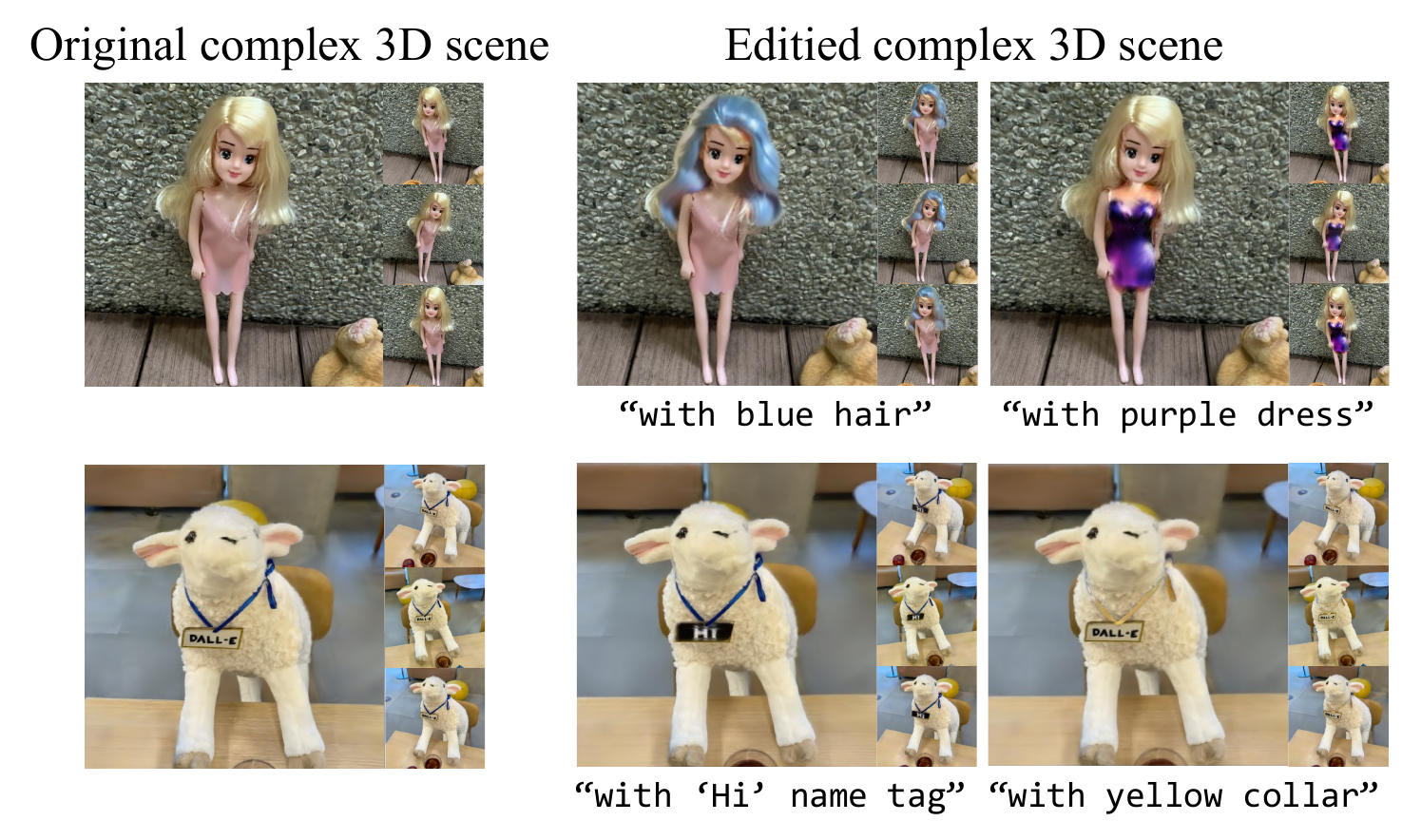}
    \caption{\textbf{3DGS part editing results in complex scenes.} We demonstrate RoMaP editing results on complex 3D Gaussian Splatting (3DGS) scenes from both the 3D-OVS and LERF datasets. As shown above, RoMaP achieves high-quality normal editing, effectively handling diverse and practical edits such as `with blue hair' or `with a `Hi' name tag'. These results highlight RoMaP’s ability to generalize across various scene complexities.}
    \label{fig:comp_editing2}
\end{figure*}

\subsubsection{Metrics} 
\paragraph{CLIP and CLIP directional score} The CLIP-based metrics calculate the cosine similarity between text and image features extracted using CLIP~\cite{radford2021learning}. CLIP scores are commonly utilized in evaluating text-to-3D~\cite{poole2022dreamfusion,tsalicoglou2024textmesh,li2024focaldreamer}. CLIP directional scores are specifically employed to evaluate whether the changes occurred in the desired direction, first introduced by~\cite{gal2022stylegan} and adopted mostly by editing models~\cite{chen2024gaussianeditor, wu2024gaussctrl, chen2024dge, cheng2023progressive3d}. We used the ViT-L/14 version of the model, with images cropped to 512 pixels and resized to 336 pixels before being input into the model.

\paragraph{TIFA and BLIP score} While CLIP-based metrics effectively evaluate coarse similarity between image and text, they have limitations in assessing fine-grained correspondences~\cite{cheng2023progressive3d, hu2023tifa, ahmadi2023examination, huang2023t2i, shi2022emscore}. To address this, we adopted two additional evaluation metrics focused on fine-grained visual-textual alignment, based on visual question answering (VQA). The TIFA score, introduced in~\cite{hu2023tifa}, measures the faithfulness of generated image to text input by generating questions with LLaMA2~\cite{touvron2023llama}, answering with UnifiedQA-v2~\cite{khashabi2022unifiedqa}. BLIP-VQA, proposed in~\cite{huang2023t2i} breaks down a prompt into multiple questions, assigning a score based on the probability of answering `yes' to each question, leveraging the vision-language understanding and generation capabilities of BLIP~\cite{li2022blip}.

\subsubsection{Implementation details} Our method is implemented in PyTorch~\cite{paszke2019pytorch}, based on Threestudio~\cite{threestudio2023}. We employ Stable Diffusion 3~\cite{esser2024scaling}. All experiments are conducted on a single A100.

\subsection{Experimental Results}
\begin{figure}
\centering
\includegraphics[width=1.0\columnwidth]{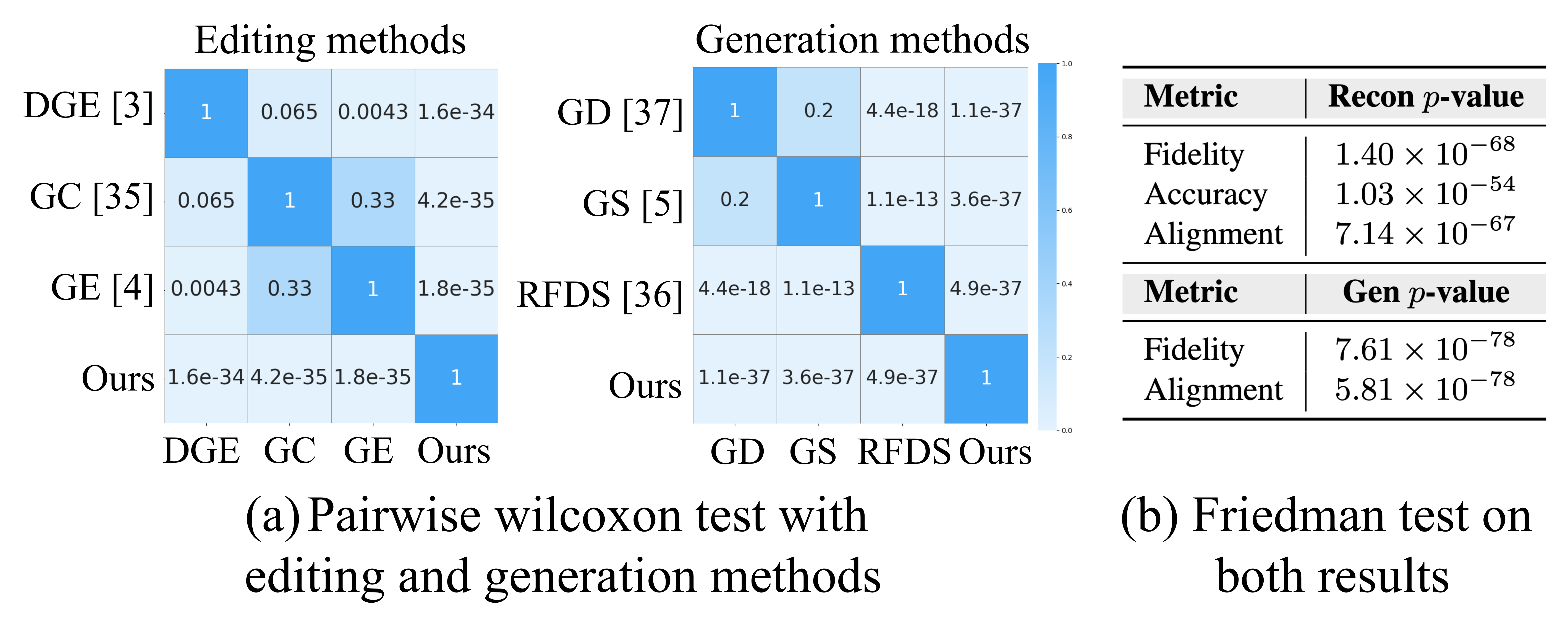}
\vspace{-0.8cm}
\caption{\textbf{Statistical results from user study.} (a) Pairwise Wilcoxon test results for editing and generation methods. (b) Friedman test p-values for fidelity, accuracy, and alignment. Our approach (Ours) achieves significantly better performance in both reconstruction and generation compared to existing methods.}
\label{fig:statistic_user}
\end{figure}

\paragraph{Quantitative results}
Detailed quantitative results are shown in Tab.~\ref{table:suptable1}, Tab.~\ref{table:suptable2}, Tab.~\ref{table:suptable3} and Tab.~\ref{table:suptable4}. The tables present quantiative results for each part editing. Our approach outperformed all other baselines in NeRF and Gaussian Splatting editing across all parts and metrics~\cite{haque2023instruct, Koo:2024PDS, dong2023vica, wu2024gaussctrl, chen2024dge, chen2024gaussianeditor}. Notably, considering that our models achieve strong performance on both CLIP-based and VQA-based scores, we can conclude that our models perform well in editing at both coarse and fine levels. Detailed results of user study for each evaluation criterion are provided in Table.~\ref{tab:recon_userstudy} and Table.~\ref{tab:gen_userstudy}. Validity of the user study result is evaluated using pairwise Wilcoxon tests and the Friedman test, as shown in Fig.\ref{fig:statistic_user}. The test results confirm that our method significantly outperforms other editing and generation methods with strong statistical significance and validating the effectiveness of our method.

\paragraph{Qualitative results}

\begin{figure*}
    \centering
     \includegraphics[width=0.78 \textwidth]{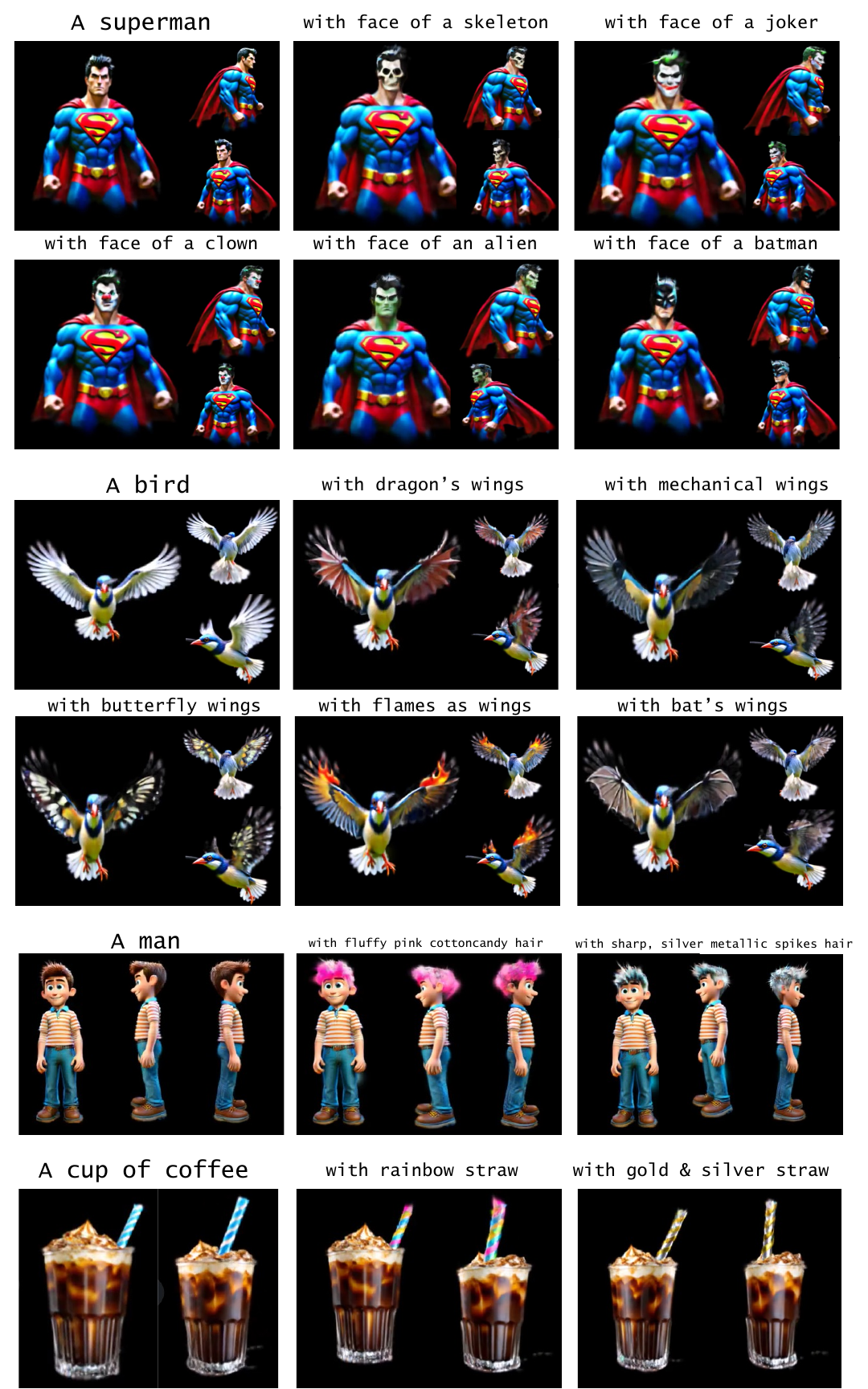}
    \caption{\textbf{Additional qualitative results of RoMaP.} Our approach, RoMaP, enables editing across a wide range of  parts, objects, and prompts in generated 3D Gaussians, further providing users with enhanced controllability over 3D content generation.}
    \label{fig:suppl_gen}
\end{figure*}

\begin{figure*}
    \centering
     \includegraphics[width=1.0 \textwidth]{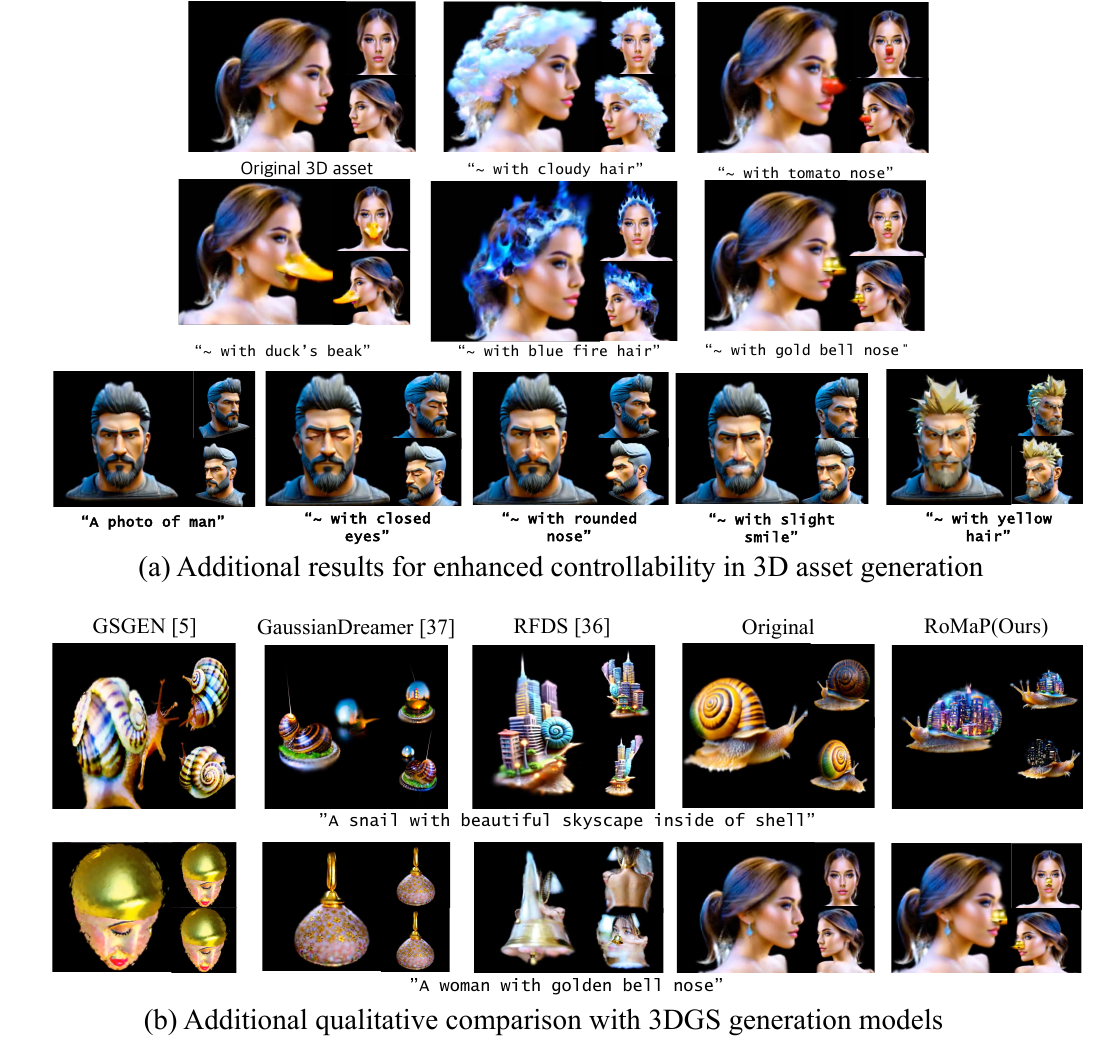}
     \vspace{-0.8cm}
    \caption{\textbf{Additional qualitative results of RoMaP.} Our approach, RoMaP, enables editing across a wide range of parts, objects, and prompts in generated 3D Gaussians, further providing users with enhanced controllability over 3D content generation.}
    \label{fig:suppl_con1}
\end{figure*}

\begin{figure*}
    \centering
     \includegraphics[width=1.0 \textwidth]{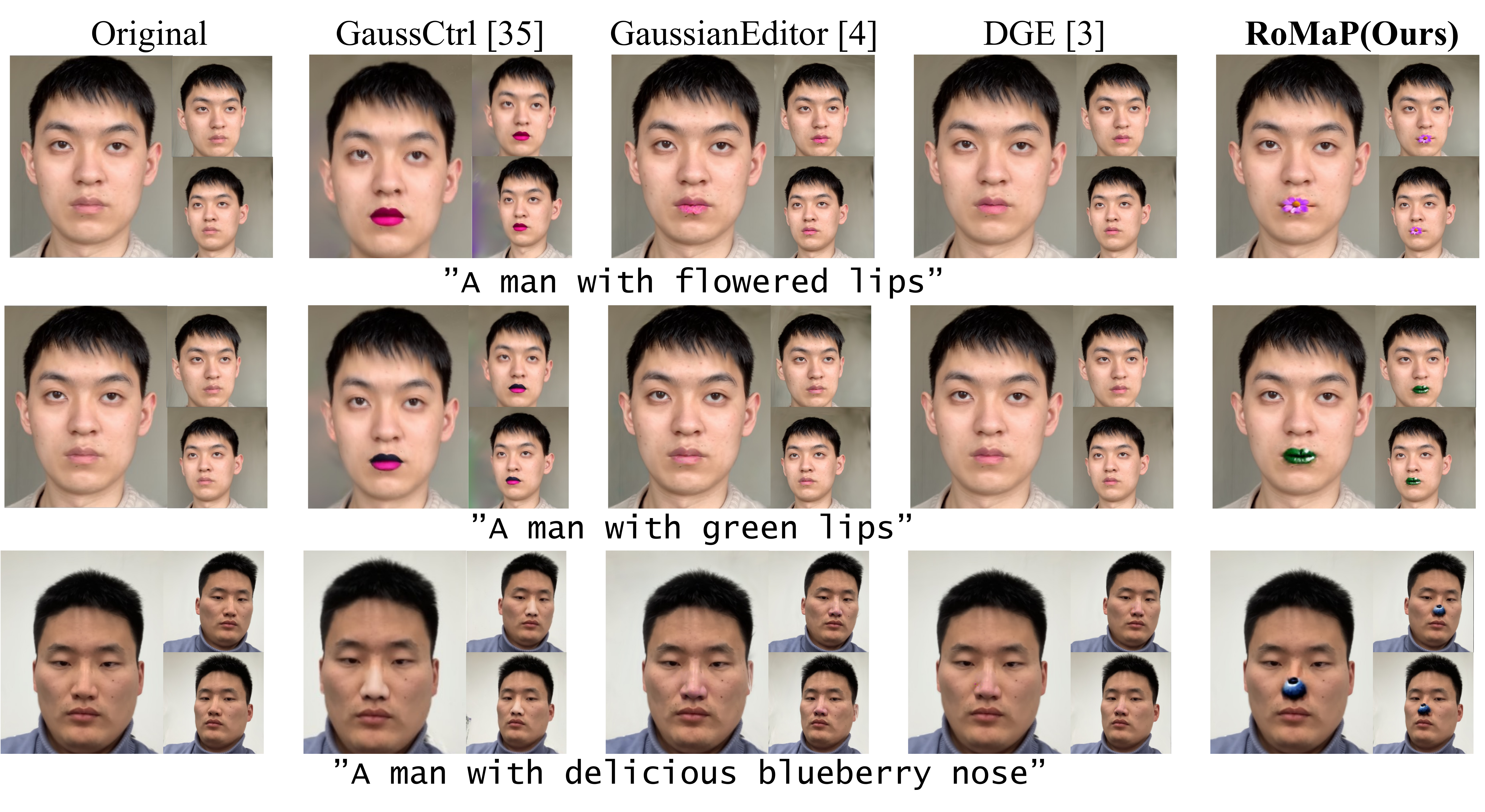}
     \vspace{-0.8cm}
    \caption{\textbf{Additional comparison results of RoMaP.} Our approach, RoMaP, enables editing across a wide range of parts, objects, compare to other methods in 3D scene reconstruction settings.}
    \label{fig:suppl_con2}
\end{figure*}

We included more qualitative results of our approach in Fig.~\ref{fig:comp_editing1}, Fig.~\ref{fig:comp_editing2}, Fig.~\ref{fig:suppl_gen}, Fig.~\ref{fig:suppl_con1}, and Fig.~\ref{fig:suppl_con2}.  As shown in Fig.~\ref{fig:suppl_gen}, Fig.~\ref{fig:suppl_con1} and Fig.~\ref{fig:suppl_con2}, our RoMaP can generate diverse 3D assets by editing the original 3D Gaussian Splatting (3DGS). Also, Fig.~\ref{fig:comp_editing1} and Fig.~\ref{fig:comp_editing2} show part-editing of our RoMaP in complex scenes. The results demonstrate that our 3D-GALP and editing strategies achieve high precision in 3D segmentation and enable precise modifications to the targeted regions, highlighting the scalability of our method to more complex and cluttered 3D scenes.

\begin{figure*}
    \centering
     \includegraphics[width=0.8 \textwidth]{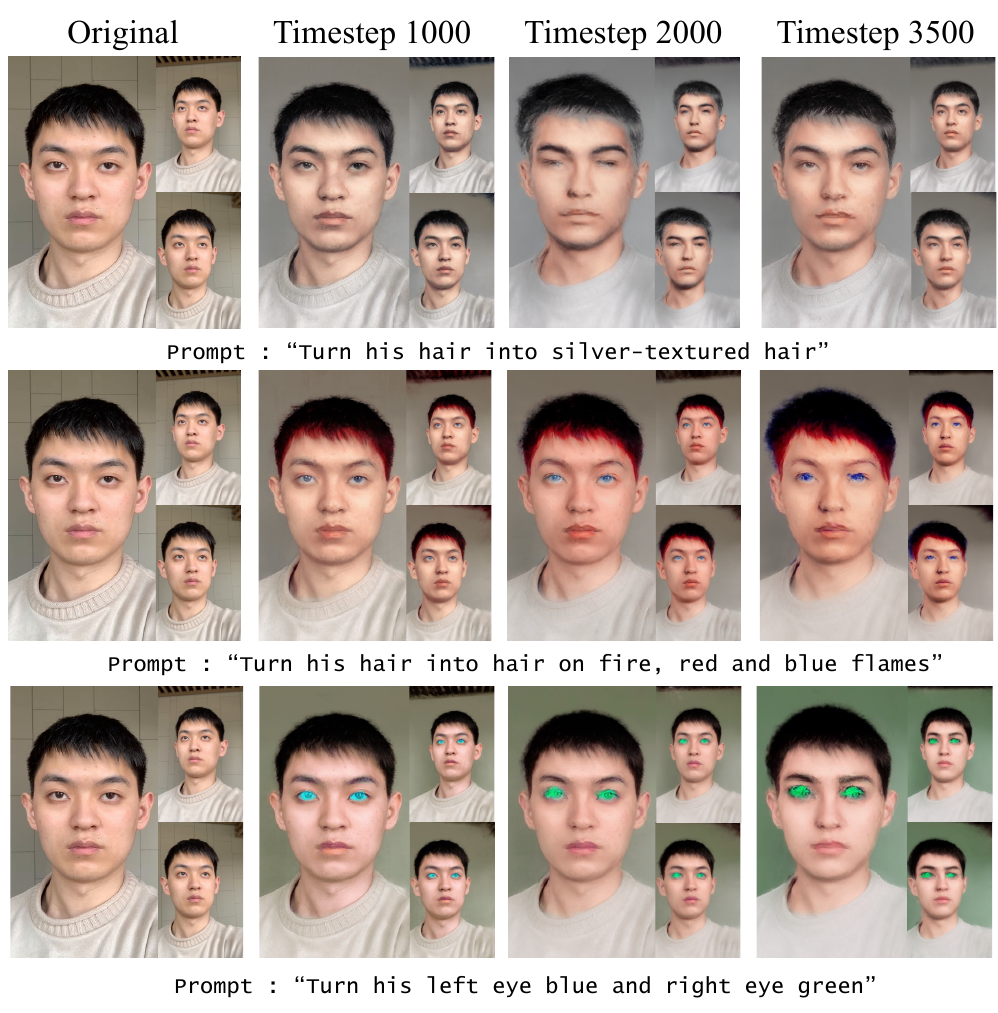}
    \caption{\textbf{Qualitative results of nerf baseines~\cite{haque2023instruct} in 3D part editing.}  }
    \label{fig:nerf_baseline}
\end{figure*}

\begin{figure*}
    \centering
     \includegraphics[width=0.75 \textwidth]{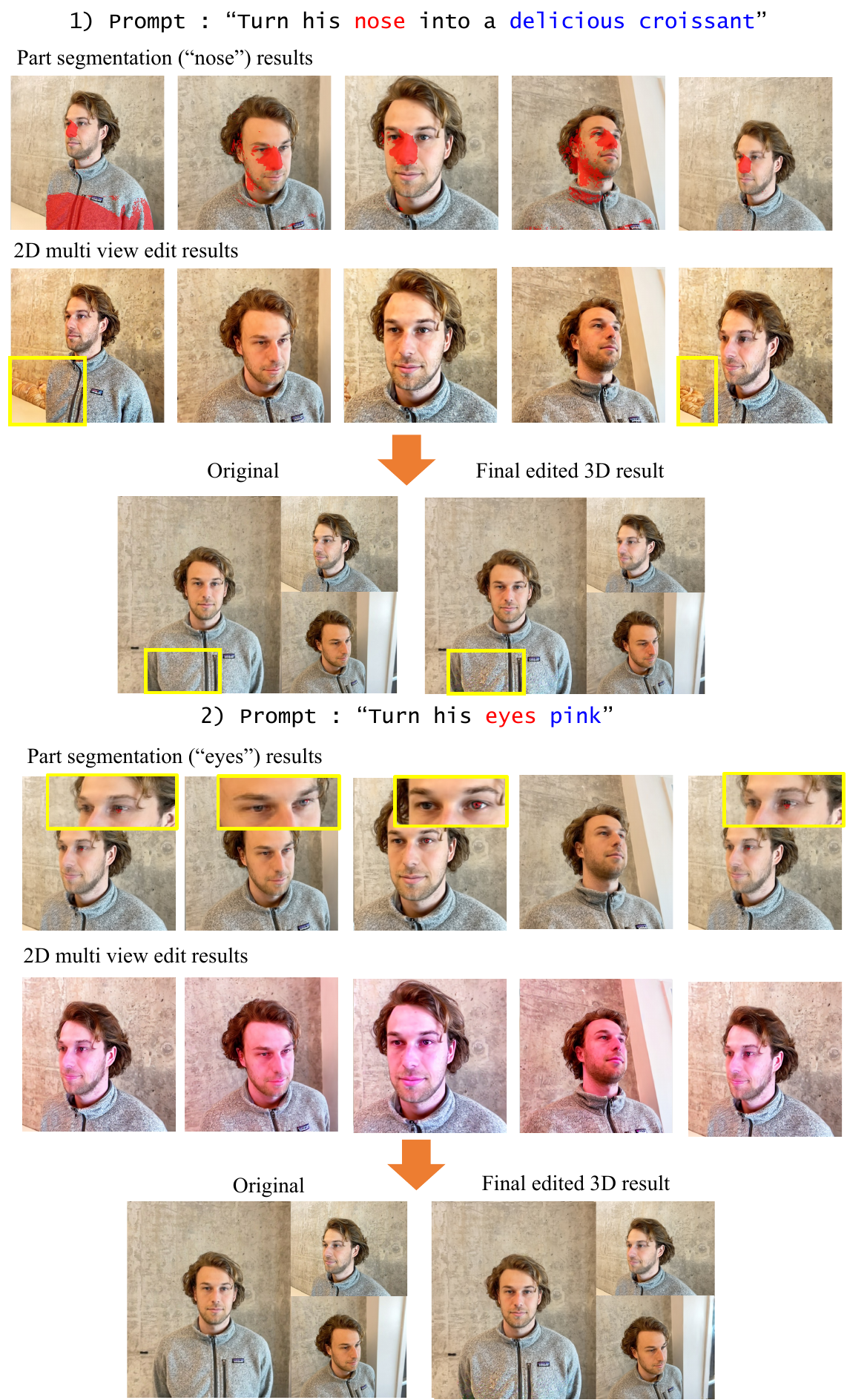}
    \caption{\textbf{Qualitative results of 3DGS baseine~\cite{chen2024gaussianeditor} in 3D part editing.}  }
    \label{fig:gaussian_baseline}
\end{figure*}

\paragraph{Qualitative results of baselines}
We visualized qualitative results of Gaussian and NeRF-editing baselines in Fig.~\ref{fig:nerf_baseline} and Fig.~\ref{fig:gaussian_baseline}. For the NeRF baseline model, we present result from IN2N~\cite{haque2023instruct}. Due to the implicit nature of NeRF, precisely selecting the target region is challenging, often resulting in unintended global changes. For example, when applying the prompt `Turn his hair into silver-textured hair', the entire scene shifts to a silver hue~\ref{fig:nerf_baseline}. Similarly, prompts such as `hair on fire' or `left eye blue and right eye green' lead to incorrect region selection, causing widespread color alterations across the scene. For the Gaussian Splatting baseline, we show results from GaussianEditor~\cite{chen2024gaussianeditor}. Inconsistencies in 2D part segmentation lead to unreliable 3D part segmentation, as shown in Fig.~\ref{fig:gaussian_baseline}. Additionally, 2D editing results demonstrate difficulties in precisely modifying the desired regions. For instance, a croissant appears in the background instead of the intended edit, or the entire scene turns pink rather than just his eyes.

\section{Additional results in complex scene}

To further validate the robustness and generalizability of RoMaP, we present additional editing results on complex 3DGS scenes from both the 3D-OVS~\cite{liu2023weakly} and LERF~\cite{kerr2023lerf} datasets. These scenes contain multiple objects with intricate part-level structures and diverse contextual settings.

As illustrated in Fig.~\ref{fig:comp_editing1}, RoMaP demonstrates precise open-vocabulary part segmentation and editing across a wide range of object types and part granularity. Examples include edits guided by prompts such as a `white cup with pink handle', `a rubber duck with white hat', and `a dog figurine with yellow eyes'. RoMaP effectively identifies and modifies fine-grained parts such as handles, beaks, collars, and ears, even under cluttered backgrounds and occlusions.

In addition, Fig.~\ref{fig:comp_editing2} further showcases our model's ability to perform practical part editing tasks involving realistic human and animal figures. Prompts such as `with blue hair', `with purple dress', and `with `Hi' name tag' illustrate RoMaP’s capability to generalize beyond common categories and execute attribute-level modifications across highly complex scenes. These results collectively highlight RoMaP's strength in both semantic understanding and fine-grained spatial localization, making it a versatile tool for open-vocabulary 3D scene editing.

\begin{figure*}
    \centering
     \includegraphics[width=0.8 \textwidth]{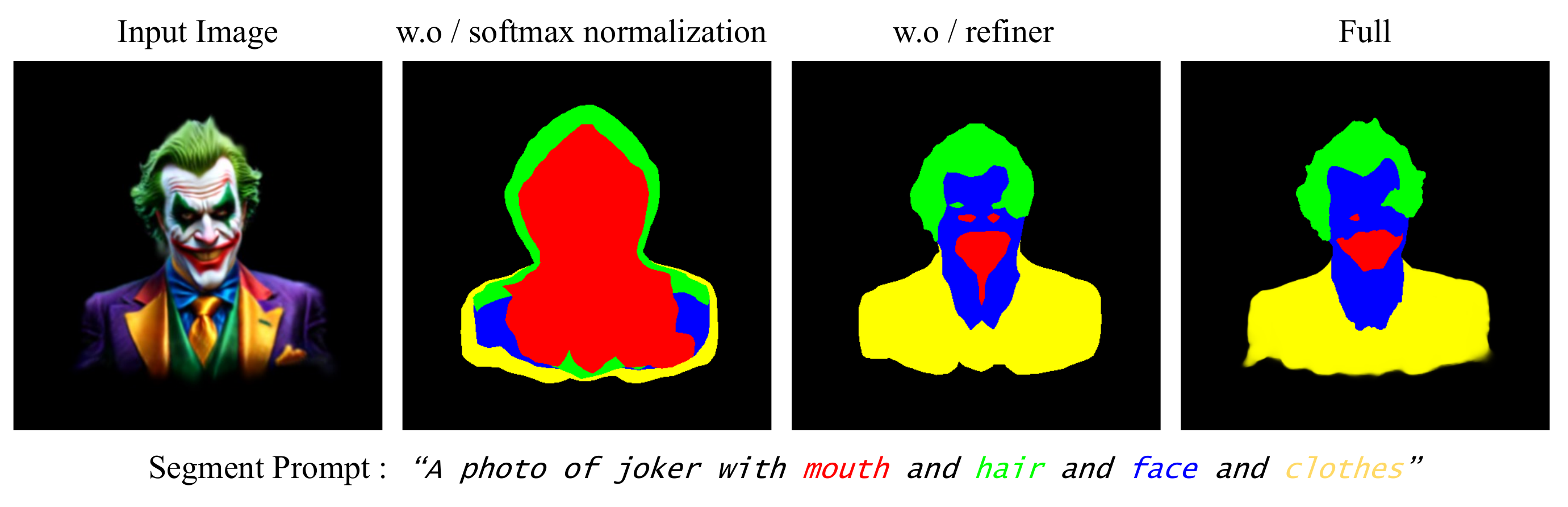}
    \caption{\textbf{Ablation study of attention map post-processing procedure} By adjusting the softmax temperature, we achieved segmentation with varying levels of granularity, while the refiner, leveraging the original image features, facilitated the segmentation of parts with sharper and more defined edges.}
    \label{fig:attn_abl}
\end{figure*}
\section{Additional validation and details of pipeline}
\label{sec:pipeline_des}
\subsection{Attention map extraction}
Unlike the naive reverse flow-matching process used in text-to-3D generation, we adopted a controlled forward ODE to extract more accurate attention maps for real images, thereby enhancing robustness. Controlled forward ODE, proposed in~\cite{rout2024semantic}, helps maintain consistency with the given image while aligning with the distribution of typical images. This balancing mechanism allows for effective inversion and editing across various inputs, especially real images, even when the given image is corrupted or atypical. 
Additionally, we adopted the approach proposed in~\cite{wang2025sclip} for dense prediction. This method allows for faster and more accurate extraction of attention maps.
\paragraph{Post-processing} We post-processed extracted attention maps by normalizing them with a softmax temperature and utilizing a refiner~\cite{cheng2020cascadepsp}. Adjusting softmax temperature allowed us to segment regions with varying granularity, while the refiner, by incorporating the original image features, enabled segmentation of parts with more precise edges, as shown in Fig.~\ref{fig:attn_abl}.
\subsection{3D-geometry aware label prediction}
\subsubsection{Details of 3D-geometry aware label prediction}
The detailed algorithm for 3D-Geometry Aware Label Prediction (3D-GALP) is provided in Algo.~\ref{algo_uncertainty}. 3D-GALP produces high-quality 3D segmentation maps even when part segmentation maps from multiple views are noisy, by applying a neighbor consistency loss that considers the soft-label property of Gaussian segmentation. Label softness is typically higher at part boundaries due to abrupt shape changes, which can lead to substantial variation in segmentation results across different views. Moreover, in practice, the Gaussians at these part boundaries may simultaneously represent pixels belonging to multiple parts depending on the viewpoint, further complicating consistent segmentation. To address this, Gaussians with both high and low softness are sampled, enabling continuous refinement of ambiguous as well as more view-invariant regions while taking surrounding information into account.

\subsubsection{Part segmentation performance of 3D-GALP compared with other language-embedded 3DGS model in complex scenes}
\label{part-seg_complex_scene}
\begin{figure}[!h]
    \centering
    \includegraphics[width=\columnwidth]{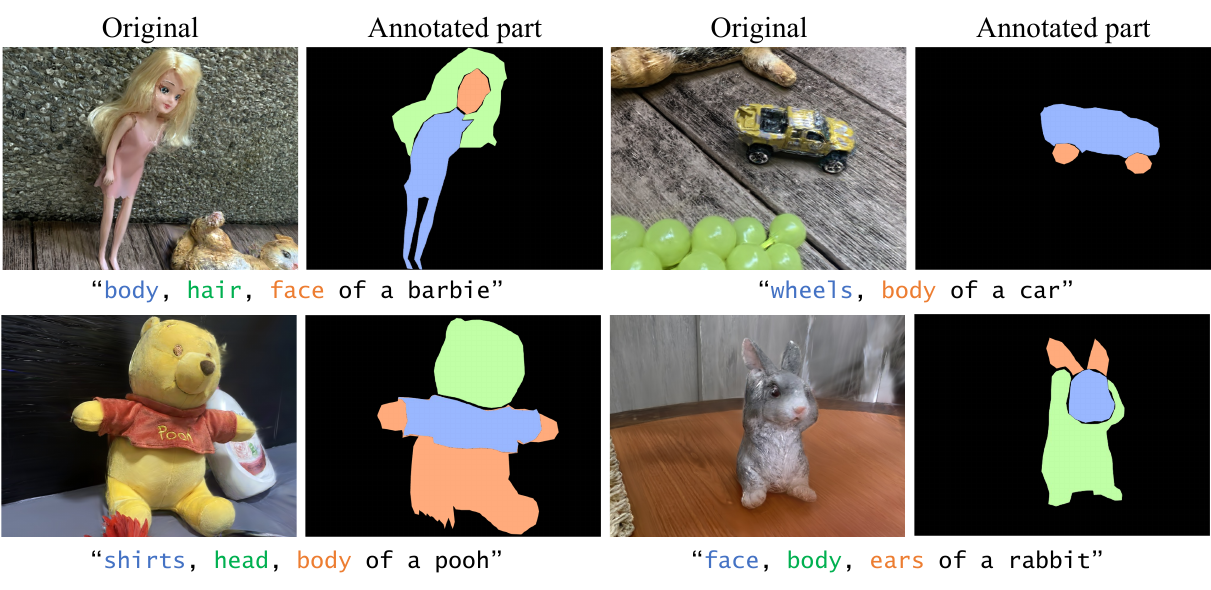}
    \caption{\textbf{Examples of part segmentation annotation in 3D-OVS dataset.}}
    \label{fig:example_annot}
\end{figure}

\begin{figure*}
    \centering
     \includegraphics[width=0.70 \textwidth]{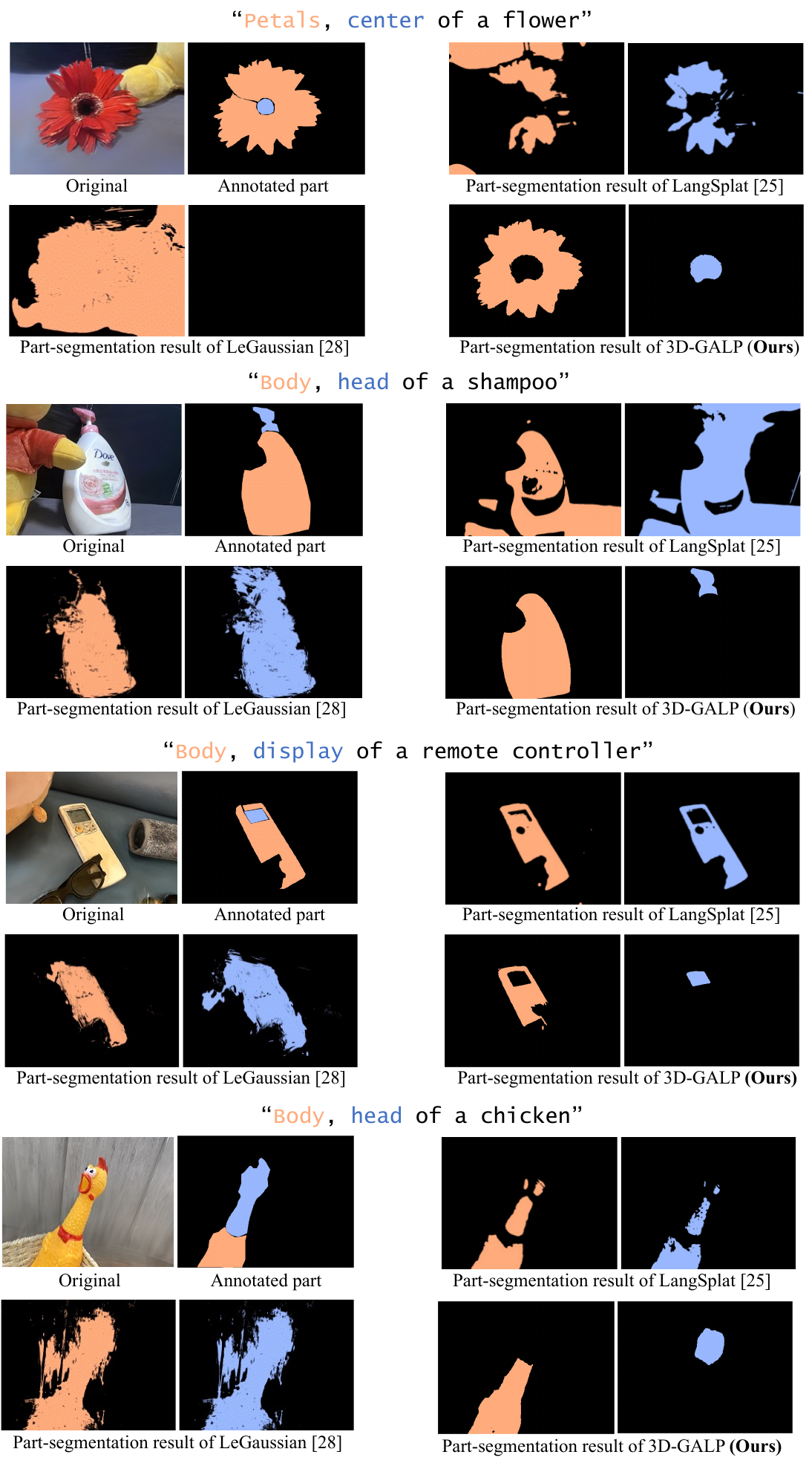}
    \caption{\textbf{Open-voca part segmentation results comparison in complicated 3DGS scenes of 3D-OVS dataset.}  }
    \label{fig:3D-OVS seg}
\end{figure*}

\paragraph{Experimental setting}
 To evaluate how effectively 3D-GALP performs part segmentation in complex scenes, we annotated part segmentation for every object in all scenes of the 3D-OVS dataset~\cite{liu2023weakly}. We compared 3D-GALP with two text-aligned segmentation models for 3D Gaussians, LangSplat~\cite{qin2024langsplat} and LeGaussian~\cite{shi2024language}. We kept hyperparameter, the softmax value for our 2D attention map extraction, to 0.2 during segmentation. We then evaluated part-segmentation results for each object from three different views, comparing them against ground truth using the mean Intersection over Union (mIoU). Examples of part-segmentation annotation are presented in Fig.~\ref{fig:example_annot}.
\paragraph{Experimental results} As shown in Tab.~\ref{tab:3dseg}, our 3D segmentation method, 3D-GALP, achieves the highest mIoU, outperforming other 3DGS segmentation baselines across all scenes. Furthermore, 3D-GALP successfully performs open-vocabulary 3DGS segmentation for parts of varying sizes in complex scenes, as illustrated in Fig.~\ref{fig:3D-OVS seg}.

\begin{table}[!h]
\centering
\resizebox{\linewidth}{!}{ 
\begin{tabular}{l|c|c|c|c|c}
\toprule
 Scene & Bench & Blue sofa & Cov.desk & Room & Average \\  \bottomrule
 LangSplat ~\cite{qin2024langsplat}& 0.005 & 0.076 & 0.093 & 0.129 & 0.076 \\  
 LeGaussian ~\cite{shi2024language} & 0.320 & 0.312 & 0.264 & 0.257 & 0.288 \\  
 \rowcolor{yellow!25} \textbf{3D-GALP (Ours)} & 
 \multicolumn{1}{c|}{\textbf{0.607}} & 
 \multicolumn{1}{c|}{\textbf{0.580}} & 
 \multicolumn{1}{c|}{\textbf{0.546}} & 
 \multicolumn{1}{c|}{\textbf{0.502}} & 
 \multicolumn{1}{c}{\textbf{0.559}} \\  
\bottomrule
\end{tabular}}
\caption{\textbf{Comparison of 3D-GALP with part segmentation on complicated 3D scenes.}}
\label{tab:3dseg}
\end{table}
\subsubsection{Ablation study on SH degree}
\begin{figure}
    \centering
     \includegraphics[width=\columnwidth]{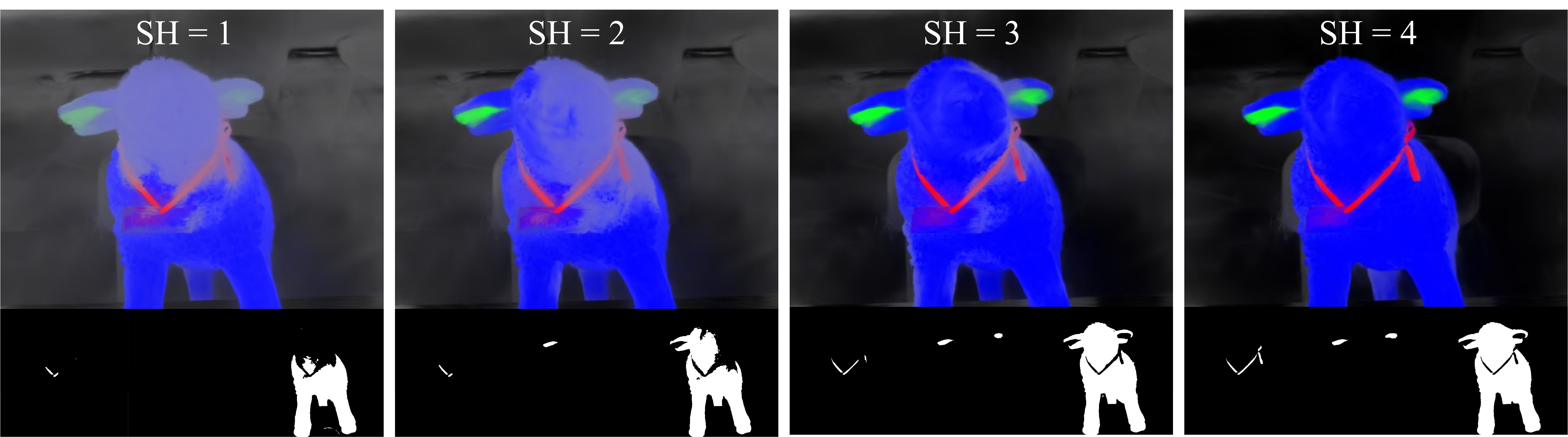}
     \vspace{-0.8cm}
    \caption{\textbf{Part-level segmentation visusalizations with different SH orders.} }
    \label{fig:sh-test}
\end{figure}

\paragraph{Experimental setting} We ablated the SH order to analyze its effect on part-level segmentation. While low-order SH is typically sufficient for modeling lighting in color representation, part-level segmentation requires sharper spatial transitions, particularly around object boundaries. To evaluate this, we conducted experiments using the same experimental settings as in~\ref{part-seg_complex_scene} with different SH degree settings.
\paragraph{Experimental results} As shown in Tab.~\ref{tab:miou_sh_transposed_avg} and Fig.~\ref{fig:sh-test}, SH=3 consistently provides the best average mIoU across scenes and captures fine-grained parts more clearly than lower orders. Although SH=4 performs best in some scenes, it introduces more noise and higher memory usage, leading to slightly worse overall performance. Based on these observations, we fix SH=3 for all segmentation experiments, as it provides the best trade-off between detail preservation and stability.
\begin{table}[!ht]
\centering
\small
\begin{tabular}{l|c|c|c|c}
\toprule
\rowcolor{gray!15}
\textbf{Order of SH} & \textbf{1} & \textbf{2} & \textbf{3} & \textbf{4} \\
\midrule
\textbf{mIoU} & 0.4777 & 0.5306 & \textbf{0.5587} & 0.5506 \\
\bottomrule
\end{tabular}
\vspace{0.2em}
\caption{\textbf{mIoU  average scores across the scenes per SH degree.} Best per scene is in \textbf{bold}.}
\label{tab:miou_sh_transposed_avg}
\end{table}
\subsection{Scheduled latent mixing and part editing}

\subsubsection{Scheduled latent mixing and part editing}
\begin{figure}
    \centering
     \includegraphics[width=0.8\columnwidth]{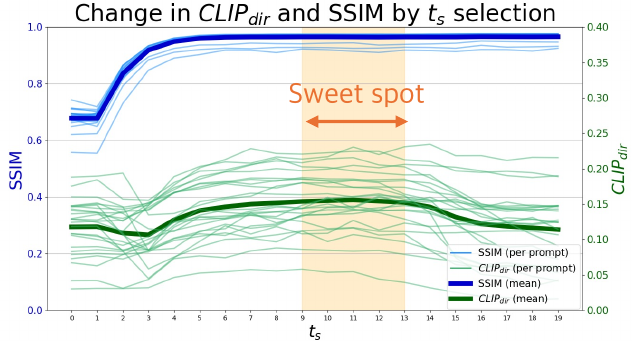}
     \vspace{-0.4cm}
    \caption{\textbf{Statistical result for finding sweet spot using CLIP and SSIM results.} }
    \label{fig:stat-clip}
\end{figure}

\begin{figure*}
    \centering
     \includegraphics[width=0.8 \textwidth]{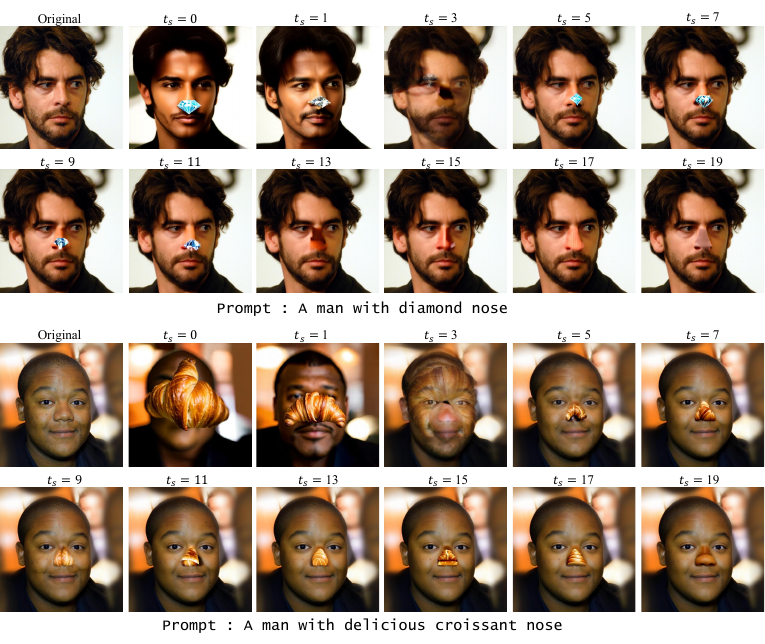}
    \caption{\textbf{Effect of different $t_s$ in SLaMP editing.}  }
    \label{fig:different_ts}
\end{figure*}

The detailed algorithm is provided in Algo.~\ref{algo_edit_inversion}. This method leverages the property of rectified flow that is more faithful to the original image. During the editing process, $\alpha_{\text{base}}$ is multiplied by the mask to ensure that regions outside the target editing area retain their original information. This introduces weak conditioning at intermediate steps of image generation, guiding the generated regions to align with the original context. At the timestep $t_s$, $\alpha_{\text{last}}$ is applied to ensure that most of the $\mathcal{M}_{\text{inv}}$ regions are replaced with $z_{\text{target}}$, preserving the majority of the reference image's information in the final output. Further results on the selection of $t_s$ are shown in Fig.~\ref{fig:different_ts}. A low $t_s$ induces dramatic changes based on the prompt, while a high $t_s$ ensures faithful adherence to the mask, taking into account the original content and its context. In the $t_s$ selection described in the main paper, we randomly selected 100 person images from the CelebAMaskHQ~\cite{lee2020maskgan} dataset, performed part-level editing using 25 prompts, and evaluated the results using CLIP$_{dir}$~\cite{gal2022stylegan} and SSIM to assess the direction of change while preserving the original content. The full experimental results with 25 prompts are shown in Fig.~\ref{fig:stat-clip}. 
 
\subsubsection{Comparison of SLaMP with other image editing models}
\paragraph{Experimental setting} To evaluate the effectiveness of our SLaMP in preserving non-target regions while accurately modifying only the specified parts compared to other models, we randomly selected 15 male and female images from the CelebAMaskHQ~\cite{lee2020maskgan} dataset. For each image, we performed image editing using 25 prompts as described in Sec.~\ref{sec:quanti_setting}. For comparison, we selected SD3-based models (SD3-inpainting~\cite{yang2024text}, Plug\&Play~\cite{esser2024scaling}, RF-inversion~\cite{rout2024semantic}), as well as an editing model based on naive latent mixing (RePaint~\cite{lugmayr2022repaint}), in contrast to our scheduled latent mixing approach. Additionally, we include a training-based model, InstructPix2Pix (IP2P~\cite{brooks2023instructpix2pix}), which is commonly adopted in 3DGS and NeRF editing approaches. For RePaint, we used a Stable Diffusion-integrated variant from HuggingFace Diffusers~\cite{von-platen-etal-2022-diffusers} library since RePaint is not originally designed for text-based image editing. We evaluated how well the changes aligned with the prompts using the CLIP$_{dir}$~\cite{gal2022stylegan} and B-VQA~\cite{huang2023t2i} metrics. 

\paragraph{Experimental results}
\begin{table}[!h]
\centering
\resizebox{\linewidth}{!}{
\begin{tabular}{l|c|c|c|c|c|c}
\toprule
\textbf{Metrics} 
& \textbf{RePaint} 
& \textbf{iP2P} 
& \textbf{SD3-inp.} 
& \textbf{Plug\&Play} 
& \textbf{RF-inv.} 
& \cellcolor{yellow!25}\textbf{SLaMP} \\
& \scriptsize\cite{lugmayr2022repaint}
& \scriptsize\cite{brooks2023instructpix2pix}
& \scriptsize\cite{esser2024scaling}
& \scriptsize\cite{yang2024text}
& \scriptsize\cite{rout2024semantic}
& \cellcolor{yellow!25}\textbf{(Ours) }\\
\midrule
CLIP$_{dir}$ $\uparrow$ & 0.111 & 0.117 & 0.147 & 0.044 & 0.089 & \cellcolor{yellow!25}\textbf{0.165} \\
B-VQA $\uparrow$        & 0.439 & 0.668 & 0.693 & 0.564 & 0.740 & \cellcolor{yellow!25}\textbf{0.758} \\
\bottomrule
\end{tabular}
}
\caption{\textbf{Quantitative comparison of SLaMP with other 2D part editing baselines.}}
\label{tab:2dedit}
\end{table}

\begin{figure*}
    \centering
     \includegraphics[width=0.8 \textwidth]{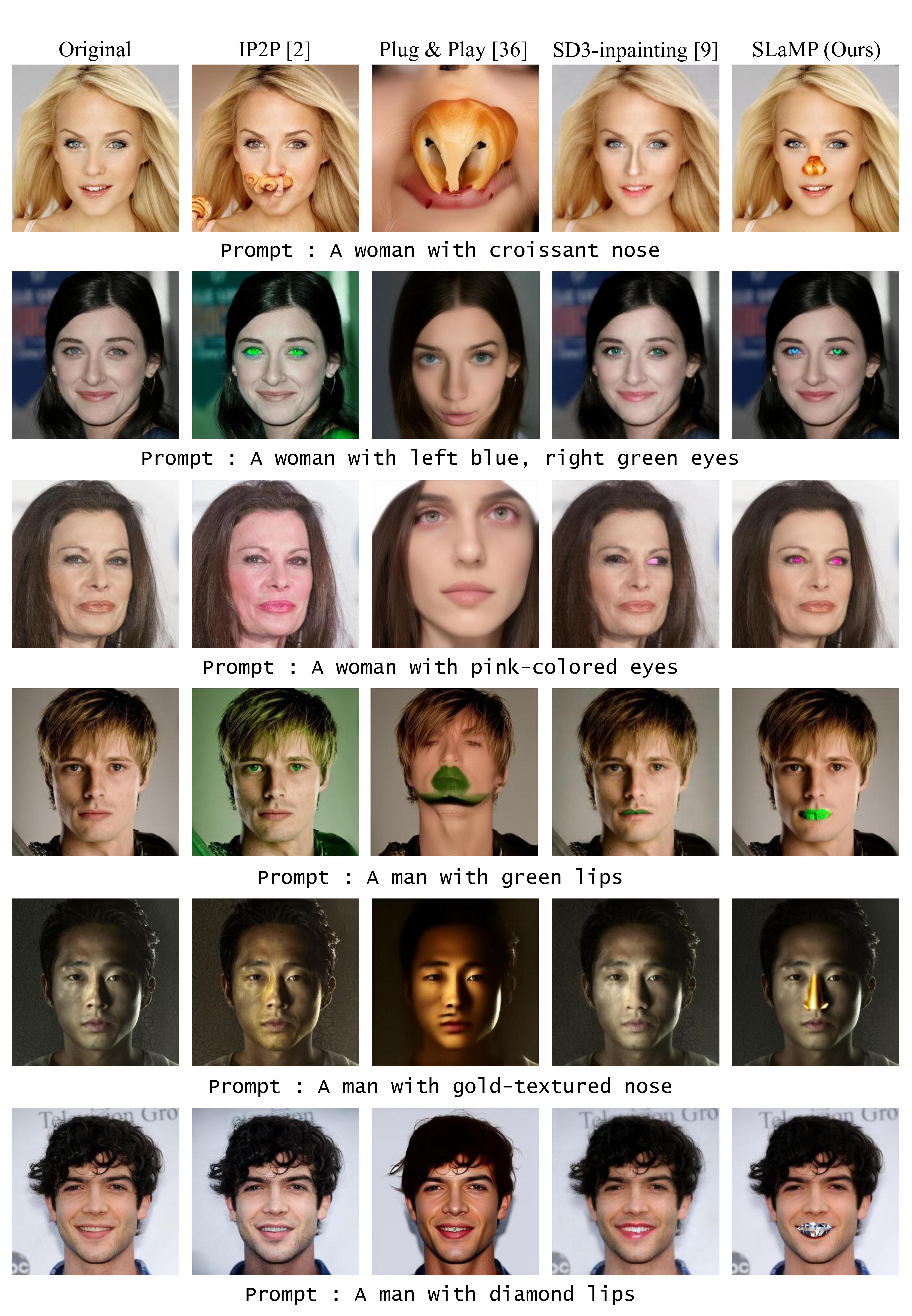}
    \caption{\textbf{Local editing results between SLaMP and 2D image editing methods.} SLaMP editing employs rectified flow inversion to achieve effective modifications while maintaining the original context in unedited regions. This contrasts with 2D image editing baselines, which struggle to edit the specified part in alignment with the text prompt.}
    \label{fig:spo_comp}
\end{figure*}

The quantitative experimental results are presented in Tab.~\ref{tab:2dedit}, and the qualitative results in Fig.~\ref{fig:spo_comp}. SLaMP outperforms all other 2D image editing baselines across all metrics, including CLIP$_{dir}$~\cite{gal2022stylegan} and BLIP-VQA~\cite{huang2023t2i}. Unlike baselines that either fail to reflect the prompt or fail to preserve the original context, SLaMP produces significant changes in the target part while accurately maintaining the untouched regions, achieving strong alignment with the text prompt.

\begin{figure*}
    \centering
     \includegraphics[width=0.8 \textwidth]{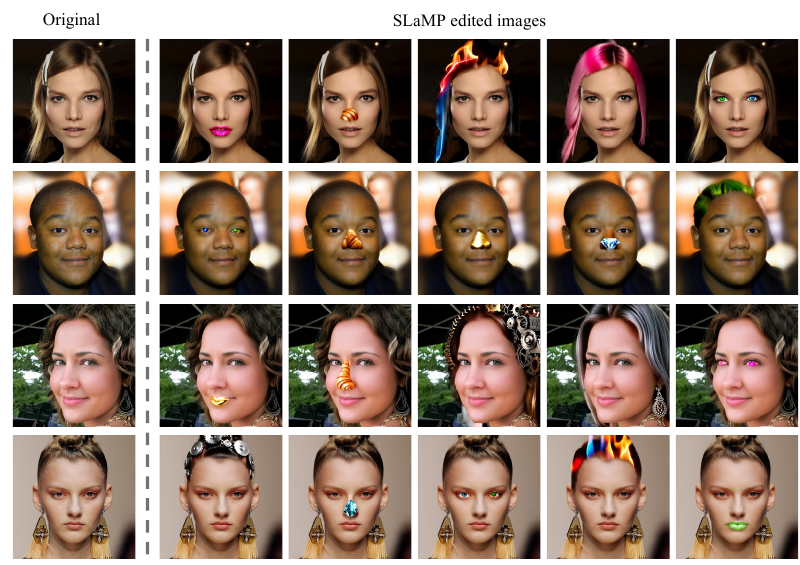}
    \caption{\textbf{More 2D part editing results with SLaMP.}}
    \label{fig:spo_results}
\end{figure*}

As shown in Fig.~\ref{fig:spo_comp}, the widely used 2D image editing baseline for 3D editing research, iP2P~\cite{brooks2023instructpix2pix}, struggles to perform meaningful part edits and often deviates from the original image context. This helps explain why existing 3D editing models often produce no visible changes in part editing tasks. RePaint~\cite{lugmayr2022repaint} employs a fixed blending ratio for harmonized inpainting, making it unsuitable for strong, prompt-driven part-level edits. In contrast, SLaMP adopts a scheduled blending strategy that enables bold edits early on and gradually preserves global context, achieving both precise modifications and faithful preservation. Additional results of SLaMP editing can be found in Fig.~\ref{fig:spo_results}.


\section{Social Impact and Limitations}
In our methodology, we utilized existing datasets from prior works~\cite{brooks2023instructpix2pix, wang2023nerf}. These datasets include information about real individuals, and if the results of our editing approach are misused, it could lead to concerns regarding negative societal impacts. Therefore, we strongly advocate for the responsible use of our methodology in adherence to ethical guidelines and relevant laws.
In perspective on limitation, our approach relies on 3D segmentation based on attention maps observed from 360-degree viewpoints. Consequently, it may not perform well when dealing with objects with highly complex geometries (\textit{e.g.}, a Klein bottle), leading to unintended editing results. Additionally, if the Gaussian Splatting scene is inherently blurry or poorly reconstructed, it becomes difficult to distinguish individual components. This can cause SD3 to fail in accurately interpreting the scene, resulting in incorrect 3D segmentation or undesired editing outcomes.

\clearpage

\renewcommand{\thealgocf}{1}

\begin{algorithm}[h!]

    \caption{3D-Geometry aware label prediction Algorithm}
    \label{algo_uncertainty}
    \DontPrintSemicolon
    \SetAlgoNoLine
    \SetAlgoVlined
    \KwIn{Gaussian Representation $\Omega$, Camera Parameters $\mathcal{C}$, Number of Anchors $K$, Nearest Neighbors $k$, Segmentation Labels $\mathbf{s}_{labels}$}
    \KwOut{Segmentation Loss $\mathcal{L}_{\text{3D}}$}

    \tcp{Initialize multi-view camera dataset}
    $\mathcal{D}_{\text{test}} \gets \text{LoadMultiviewDataset}(\mathcal{C})$

    \tcp{Compute SH consistency}
    $\mathbf{S} \gets \Omega.\text{get\_sh\_objects}()$ \;
    $\mathbf{T} \gets \emptyset$ \tcp*{Store SH values for different views}

    \ForEach{$\mathbf{b}$ in $\mathcal{D}_{\text{test}}$}{
        $\mathbf{d} \gets \text{ComputeViewDirection}(\mathbf{b}, \mathcal{C})$ \;
        $\mathbf{s}_b \gets \text{EvalSH}(\Omega, \mathbf{S}, \mathbf{d})$ \;
        $\mathbf{T} \gets \mathbf{T} \cup \mathbf{s}_b$ \;
    }

    \tcp{Compute variance and entropy for each Gaussian}
    \ForEach{Gaussian $i$ in $\Omega$}{
        Compute variance:  
        $\boldsymbol{v_i} \gets \frac{1}{|\mathbf{T}|} \sum_{\mathbf{r} \in \mathbf{T}} \|\mathbf{r} - \bar{\mathbf{r}}\|^2$,
        where $\bar{\mathbf{r}} = \frac{1}{|\mathbf{T}|} \sum_{\mathbf{r} \in \mathbf{T}} \mathbf{r}$

        Compute entropy:  
        $\mathbf{sim} \gets \frac{\bar{\mathbf{r}} \cdot \mathbf{R}_{\text{labels}}}{\|\bar{\mathbf{r}}\| \|\mathbf{R}_{\text{labels}}\|}$ \;
        $\mathbf{p_i} \gets \frac{e^{\mathbf{sim}}}{\sum e^{\mathbf{sim}}}$ \;
        $\mathbf{H_i} \gets -\sum \mathbf{p_i} \log (\mathbf{p_i} + \epsilon)$ \tcp*{Compute entropy}

        Compute label softness:  
        $\mathbf{U_i} \gets \mathbf{H_i} \cdot \boldsymbol{v_i}$ \;
    }

    \tcp{Anchor Selection Based on label softness}
    Sort all Gaussians by $U_i$ in descending order \;
    Select $\lfloor K/2 \rfloor$ anchors with highest $U_i$ \;
    Select $\lfloor K/2 \rfloor$ anchors with lowest $U_i$ \;
    Define set of selected anchors: $S$ \;

    \tcp{Compute Anchor-Based Neighbor Consistency Loss}
    \ForEach{anchor $i \in S$}{
        Find nearest neighbors $\mathcal{N}_k(i) = \{ j_1,\dots,j_k \}$ using Euclidean distance \;
        
        Compute L1 loss:  
        $\mathcal{L}_{\text{3D}} \gets \sum_{i \in S} \left[ \frac{1}{k} \sum_{j \in \mathcal{N}_k(i)} \| \mathbf{r}_i - \mathbf{r}_j \|_1 \right]$ \;
    }

    \Return{$\mathcal{L}_{\text{3D}}$}
\caption{\textbf{Algorithm of 3D-geometry aware label prediction (3D-GALP).}  }
\end{algorithm}

\renewcommand{\thealgocf}{2}

\begin{algorithm}[h!]
    \caption{Scheduled latent mixing and part editing Algorithm}
    \label{algo_edit_inversion}
    \DontPrintSemicolon
    \SetAlgoNoLine
    \SetAlgoVlined
    \KwIn{Latents $\mathbf{z}$, Text Embeddings $\mathbf{E}$, Camera Condition $\mathbf{C}$, Timestep $\mathbf{T}$, Noise $\mathbf{n}_{\text{target}}$,   Cfg scale c, $\gamma$, $\eta_{\text{values}}$, $\alpha_{base}$, $\alpha_{last}$, Mask $\mathcal{M}$, , Mix timestep $t_s$ }
    \KwOut{Model Prediction $\mathbf{m}_{\text{pred}}$}

    \tcp{Latent Initialization and Noise Target}
    
    \For{$t_{\text{curr}}, t_{\text{prev}}$ in $\text{timesteps}[:-1], \text{timesteps}[1:]$}{
        $\mathbf{t} \gets t_{\text{curr}} \times 1000$ \;
        $\mathbf{v}_{\text{pred}} \gets \text{transformer}(\mathbf{z}_{\text{noisy}}, \mathbf{t}, \mathbf{E}_{\text{uncond}})$ \;
        $\mathbf{v}_{\text{target}} \gets (\mathbf{n}_{\text{target}} - \mathbf{z}_{\text{noisy}}) / (1 - t_{\text{curr}})$ \;
        $\mathbf{v}_{\text{interp}} \gets \gamma \cdot \mathbf{v}_{\text{target}} + (1 - \gamma) \cdot \mathbf{v}_{\text{pred}}$ \;
        $\mathbf{z}_{\text{noisy}} \gets \mathbf{z}_{\text{noisy}} + (t_{\text{prev}} - t_{\text{curr}}) \cdot \mathbf{v}_{\text{interp}}$ \;
    }
    
    $\mathbf{z}_{\text{target}} \gets \mathbf{z}.\text{clone}$ \;
    \For{$t$ in $\text{timesteps}$}{
        $\mathbf{t} \gets t / 1000$ \;
        $\mathbf{v}_{\text{pred}} \gets \text{transformer}(\mathbf{z}_{\text{noisy}}, \mathbf{t}, \mathbf{E}_{\text{mix}})$ \;
        $\mathbf{v}_{\text{target}} \gets -(\mathbf{z}_{\text{target}} - \mathbf{z}_{\text{noisy}}) / t$ \;
        $\eta \gets \eta_{\text{values}}[i]$ \;
        $\mathbf{v}_{\text{interp}} \gets \mathbf{v}_{\text{pred}} + \eta \cdot (\mathbf{v}_{\text{target}} - \mathbf{v}_{\text{pred}})$ \;
        $\mathbf{z}_{\text{noisy}} \gets \text{scheduler.step}(\mathbf{v}_{\text{interp}}, t, \mathbf{z}_{\text{noisy}})$ \;
        $\mathcal{F} \gets  \alpha_{last}$ if $i > |\text{timesteps}| - t_s$ else  $\alpha_{base}$ \;
        $\mathcal{M}_{inv} \gets \mathcal{F} \times ( 1-\mathcal{M} )$
        $\mathbf{z}_{\text{noisy}} \gets \mathbf{z}_{\text{noisy}} \times (1 - \mathcal{M}_{inv}) + \mathbf{z}_{\text{target}} \times \mathcal{M}_{inv}$ \;
    }
    $\mathbf{m}_{\text{pred}} \gets \mathbf{z}_{\text{noisy}}$ \;
    \Return{$\mathbf{m}_{\text{pred}}$}
\end{algorithm}


\clearpage
\small
\bibliographystyle{ieeenat_fullname}
\bibliography{main}


\end{document}